%% file: main.tex
\documentclass{article}

\usepackage[final]{corl_2020} %

\usepackage{microtype}
\usepackage{graphicx}
\usepackage{subfigure}
\usepackage{hyperref}
\usepackage[utf8]{inputenc} %
\usepackage[T1]{fontenc}    %
\usepackage{url}            %
\usepackage{booktabs}       %
\usepackage{amsfonts}       %
\usepackage{nicefrac}       %
\usepackage{color}
\usepackage{mathtools}
\usepackage{amsmath,amssymb}
\usepackage[svgnames]{xcolor}
\usepackage{bm}
\usepackage{siunitx}
\usepackage{wrapfig}
\usepackage{algorithm}
\usepackage[noend]{algorithmic}
\usepackage{lipsum}

\input{math_commands.tex}

\newcommand{\ie}{i.e., }
\newcommand{\eg}{e.g., }
\newcommand{\Skip}[1]{}

\newcommand{\spirl}[0]{SPiRL}

\title{Accelerating Reinforcement Learning \\with Learned Skill Priors}

\author{
  Karl Pertsch \quad\quad Youngwoon Lee \quad\quad Joseph J.~Lim\vspace{5pt}\\
  Department of Computer Science \\
  University of Southern California \\
  \texttt{\{pertsch,lee504,limjj\}@usc.edu}\\
}

\begin{document}
\maketitle

\input{sections/abstract}
\input{sections/introduction}

\input{sections/related_work}

\input{sections/approach}

\input{sections/experiments}

\input{sections/conclusion}

\clearpage
{\small
    \bibliography{bibref_definitions_long,bibtex}
}
\clearpage
\appendix
\input{sections/appendix.tex}

\end{document}

%% file: math_commands.tex
\usepackage{amsmath,amsfonts,bm}

\def\eqref#1{equation~\ref{#1}}

\def\1{\bm{1}}

\def\va{{\bm{a}}}

\DeclareMathAlphabet{\mathsfit}{\encodingdefault}{\sfdefault}{m}{sl}
\SetMathAlphabet{\mathsfit}{bold}{\encodingdefault}{\sfdefault}{bx}{n}

%% file: sections/abstract.tex
\begin{abstract}
Intelligent agents rely heavily on prior experience when learning a new task, yet most modern reinforcement learning (RL) approaches learn every task from scratch. One approach for leveraging prior knowledge is to transfer \emph{skills} learned on prior tasks to the new task. However, as the amount of prior experience increases, the number of transferable skills grows too, making it challenging to explore the full set of available skills during downstream learning. Yet, intuitively, not all skills should be explored with equal probability; for example information about the current state can hint which skills are promising to explore. In this work, we propose to implement this intuition by learning a \emph{prior over skills}. We propose a deep latent variable model that jointly learns an embedding space of skills and the skill prior from offline agent experience. We then extend common maximum-entropy RL approaches to use skill priors to guide downstream learning. We validate our approach, \spirl~(Skill-Prior RL), on complex navigation and robotic manipulation tasks and show that learned skill priors are essential for effective skill transfer from rich datasets. Videos and code are available at \href{https://clvrai.com/spirl}{\texttt{clvrai.com/spirl}}.
\end{abstract}

\keywords{Reinforcement Learning, Skill Learning, Transfer Learning} 

%% file: sections/introduction.tex
\section{Introduction}
\label{sec:intro}

Intelligent agents are able to utilize a large pool of prior experience to efficiently learn how to solve new tasks~\cite{woodworth1901influence}. %
In contrast, reinforcement learning (RL) agents typically learn each new task \emph{from scratch}, without leveraging prior experience. %
Consequently, agents need to collect a large amount of experience while learning the target task, %
which is expensive, especially in the real world. On the other hand, there is an abundance of collected agent experience available in domains like autonomous driving~\cite{nuscenes2019}, %
indoor navigation~\cite{Mo18AdobeIndoorNav}, %
or robotic manipulation~\cite{dasari2019robonet,cabi2019}. %
With the widespread deployment of robots on streets or in warehouses the available amount of data will further increase in the future. However, the majority of this data is unstructured, without clear task or reward definitions, making it difficult to use for learning new tasks. In this work, our aim is to devise a scalable approach for leveraging such unstructured experience to accelerate the learning of new downstream tasks.

One flexible way to utilize unstructured prior experience is by extracting \emph{skills}, temporally extended actions that represent useful behaviors, which can be repurposed to solve downstream tasks. Skills can be learned from data without any task or reward information and can be transferred to new tasks and even new environment configurations. %
Prior work has learned skill libraries from data collected by humans~\cite{schaal2006adaptive,merel2018neural,merel2019reusable,shankar2019discovering,lynch2020learning} or by agents autonomously exploring the world~\cite{hausman2018learning,sharma2019dynamics}. %
To solve a downstream task using the learned skills, these approaches train a high-level policy whose action space is the set of extracted skills. The dimensionality of this action space scales with the number of skills. Thus, the large skill libraries extracted from rich datasets can, somewhat paradoxically, lead to worse learning efficiency on the downstream task, since the agent needs to collect large amounts of experience to perform the necessary exploration in the space of skills~\cite{jong2008utility}. 

\begin{figure}[t]
    \centering
    \includegraphics[width=1\linewidth]{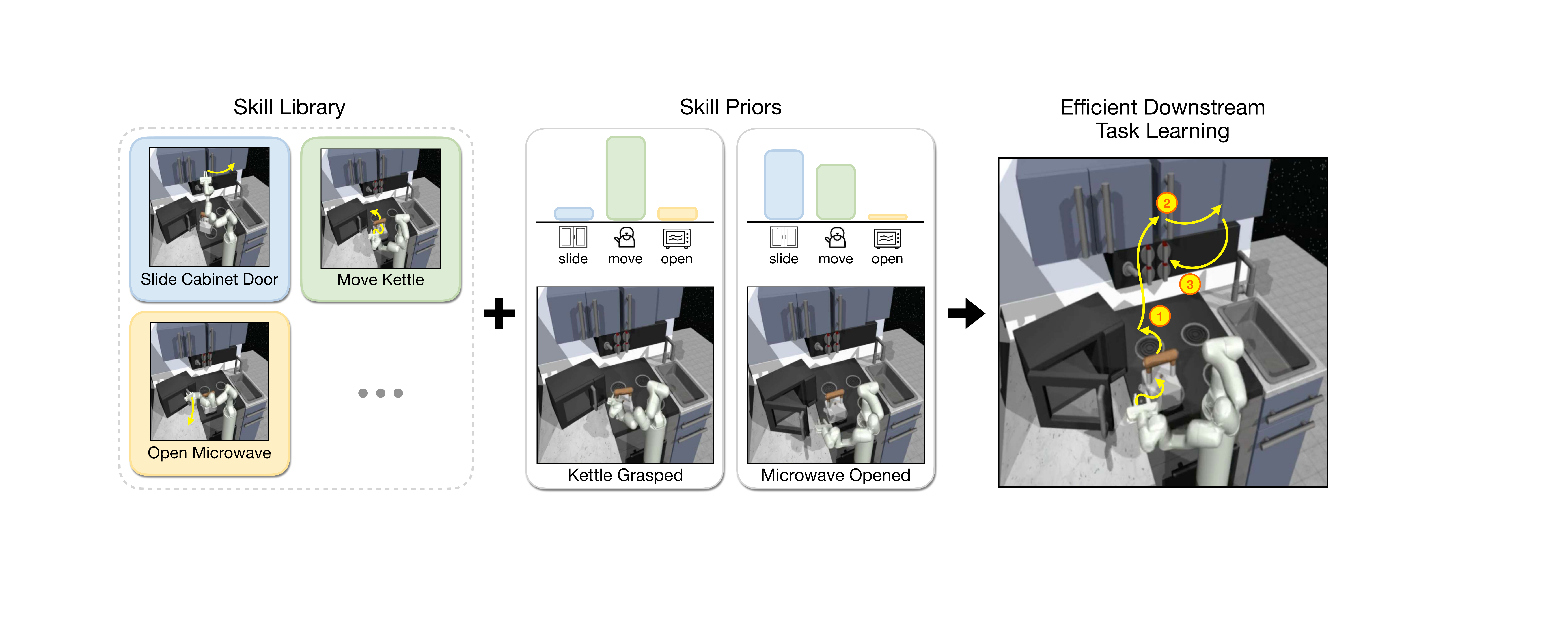}
    \caption{Intelligent agents can use a large library of acquired skills when learning new tasks. Instead of exploring skills uniformly, they can leverage \emph{priors over skills} as guidance, based \eg on the current environment state. Such priors capture which skills are promising to explore, like moving a kettle when it is already grasped, and which are less likely to lead to task success, like attempting to open an already opened microwave. In this work, we propose to jointly learn an embedding space of skills and a prior over skills from unstructured data to accelerate the learning of new tasks.
    }
    \label{fig:overview}
\end{figure}

The key idea of this work is to learn a \emph{prior over skills} along with the skill library to guide exploration in skill space and enable efficient downstream learning, even with large skill spaces. Intuitively, the prior over skills is not uniform: if the agent holds the handle of a kettle, it is more promising to explore a pick-up skill than a sweeping skill~(see Fig.~\ref{fig:overview}). To implement this idea, we design a stochastic latent variable model that learns a continuous embedding space of skills and a prior distribution over these skills from unstructured agent experience. We then show how to naturally incorporate the learned skill prior into maximum-entropy RL algorithms for efficient learning of downstream tasks. To validate the effectiveness of our approach, \spirl~(Skill-Prior RL), we apply it to complex, long-horizon navigation and robot manipulation tasks. We show that through the transfer of skills we can use unstructured experience for accelerated learning of new downstream tasks and that learned skill priors are essential to efficiently utilize rich experience datasets.

In summary, our contributions are threefold: (1)~we design a model for jointly learning an embedding space of skills and a prior over skills from unstructured data, (2)~we extend maximum-entropy RL to incorporate learned skill priors for downstream task learning, and (3)~we show that learned skill priors accelerate learning of new tasks across three simulated navigation and robot manipulation tasks.

%% file: sections/related_work.tex
\section{Related Work}
\label{sec:related_work}

The goal of our work is to leverage prior experience for accelerated learning of downstream tasks. \textbf{Meta-learning} approaches~\cite{finn2017model,rakelly2019efficient} %
similarly aim to extract useful priors from previous experience to improve the learning efficiency for unseen tasks. However, they require a defined set of training tasks and online data collection during pre-training and therefore cannot leverage large offline datasets. In contrast, our model learns skills fully offline from unstructured data.

Approaches that operate on such offline data are able to leverage large existing datasets~\cite{dasari2019robonet,cabi2019} %
and can be applied to domains where data collection is particularly costly or safety critical~\cite{levine2020offline}. A number of works have recently explored the \textbf{offline reinforcement learning} setting~\cite{levine2020offline,fujimoto2019off,jaques2019way,kumar2019stabilizing,wu2019behavior}, in which a task needs to be learned purely from logged agent experience without any environment interactions. It has also been shown how offline RL can be used to accelerate online RL~\cite{nair2020accelerating}. However, these approaches require the experience to be annotated with rewards for the downstream task, which are challenging to provide for large, real-world datasets, %
especially when the experience is collected across a wide range of tasks. Our approach based on skill extraction, on the other hand, does not require any reward annotation on the offline experience data and, once extracted, skills can be reused for learning a wide range of downstream tasks. 

\Skip{
Offline data can also be used for learning dynamics models, which can be leveraged in \textbf{model-based reinforcement learning} approaches~\cite{ebert2018visual,fang2019dynamics} %
for sample-efficient downstream task learning. However, modelling the full dynamics of complex environments is challenging and model-based approaches typically suffer from accumulating errors. Modeling comparatively low-dimensional agent behaviors in the form of skills can provide a more efficient way to leverage offline experience.
}

More generally, the problem of inter-task transfer has been studied for a long time in the RL community~\cite{taylor2009transfer}. The idea of \textbf{transferring skills between tasks} dates back at least to the \textsc{SKILLS}~\cite{thrun1995finding} and \textit{PolicyBlocks}~\cite{pickett2002policyblocks} algorithms. Learned skills can be represented as sub-policies in the form of options~\cite{sutton1999between,bacon2017option}, as subgoal setter and reacher functions~\cite{gupta2019relay,mandlekar2019iris} %
or as discrete primitive libraries~\citep{schaal2006adaptive,lee2018composing}. %
Recently, a number of works have explored the embedding of skills into a continuous \emph{skill space} via stochastic latent variable models~\citep{hausman2018learning,merel2018neural,kipf2018compositional,merel2019reusable,shankar2019discovering,whitney2019dynamics,lynch2020learning}. %
When using powerful latent variable models, these approaches are able to represent a very large number of skills in a compact embedding space. However, the exploration of such a rich skill embedding space can be challenging, leading to inefficient downstream task learning~\cite{jong2008utility}. Our work introduces a learned skill prior to guide the exploration of the skill embedding space, enabling efficient learning on rich skill spaces.

\textbf{Learned behavior priors} are commonly used to guide task learning in offline RL approaches~\cite{fujimoto2019off,jaques2019way,wu2019behavior} in order to avoid value overestimation for actions outside of the training data distribution. Recently, action priors have been used to leverage offline experience for learning downstream tasks~\cite{siegel2020keep}. Crucially, our approach learns priors over \emph{temporally extended actions} (i.e., skills) allowing it to scale to complex, long-horizon downstream tasks.

%% file: sections/approach.tex
\section{Approach}
\label{sec:approach}

Our goal is to leverage skills extracted from large, unstructured datasets to accelerate the learning of new tasks. Scaling skill transfer to large datasets is challenging, since learning the downstream task requires picking the appropriate skills from an increasingly large library of extracted skills. In this work, we propose to use learned skill priors to guide exploration in skill space and allow for efficient skill transfer from large datasets. We decompose the problem of prior-guided skill transfer into two sub-problems: (1)~the extraction of skill embedding and skill prior from offline data, and (2)~the prior-guided learning of downstream tasks with a hierarchical policy.

\subsection{Problem Formulation}
\label{sec:problem_formulation}

We assume access to a dataset $\mathcal{D}$ of pre-recorded agent experience in the form of state-action trajectories $\tau_i = \{(s_0, a_0), \dots, (s_{T_i}, a_{T_i})\}$. This data can be collected using previously trained agents across a diverse set of tasks~\cite{fu2020d4rl,gulcehre2020rl}, through agents autonomously exploring their environment~\cite{hausman2018learning,sharma2019dynamics}, %
via human teleoperation~\cite{schaal2005motion,gupta2019relay,mandlekar2018roboturk,lynch2020learning} or any combination of these. Crucially, we aim to leverage \emph{unstructured} data that does not have annotations of tasks or sub-skills and does not contain reward information to allow for scalable data collection on real world systems. %
In contrast to imitation learning problems we do not assume that the training data contains complete solutions for the downstream task. Hence, we focus on transferring skills to new problems.

The downstream learning problem is formulated as a Markov decision process (MDP) defined by a tuple $\{\mathcal{S}, \mathcal{A}, \mathcal{T}, R, \rho, \gamma\}$ of states, actions, transition probability, reward, initial state distribution, and discount factor. We aim to learn a policy $\pi_\theta(a \vert s)$ with parameters $\theta$ that maximizes the discounted sum of rewards $J(\theta) = \mathbb{E}_\pi \big[ \sum_{t=0}^{T-1} \gamma^t r_t \big]$ where $T$ is the episode horizon.

\subsection{Learning Continuous Skill Embedding and Skill Prior}
\label{sec:skill_embed_prior_learning}

\begin{figure}[t]
    \centering
    \includegraphics[width=1\linewidth]{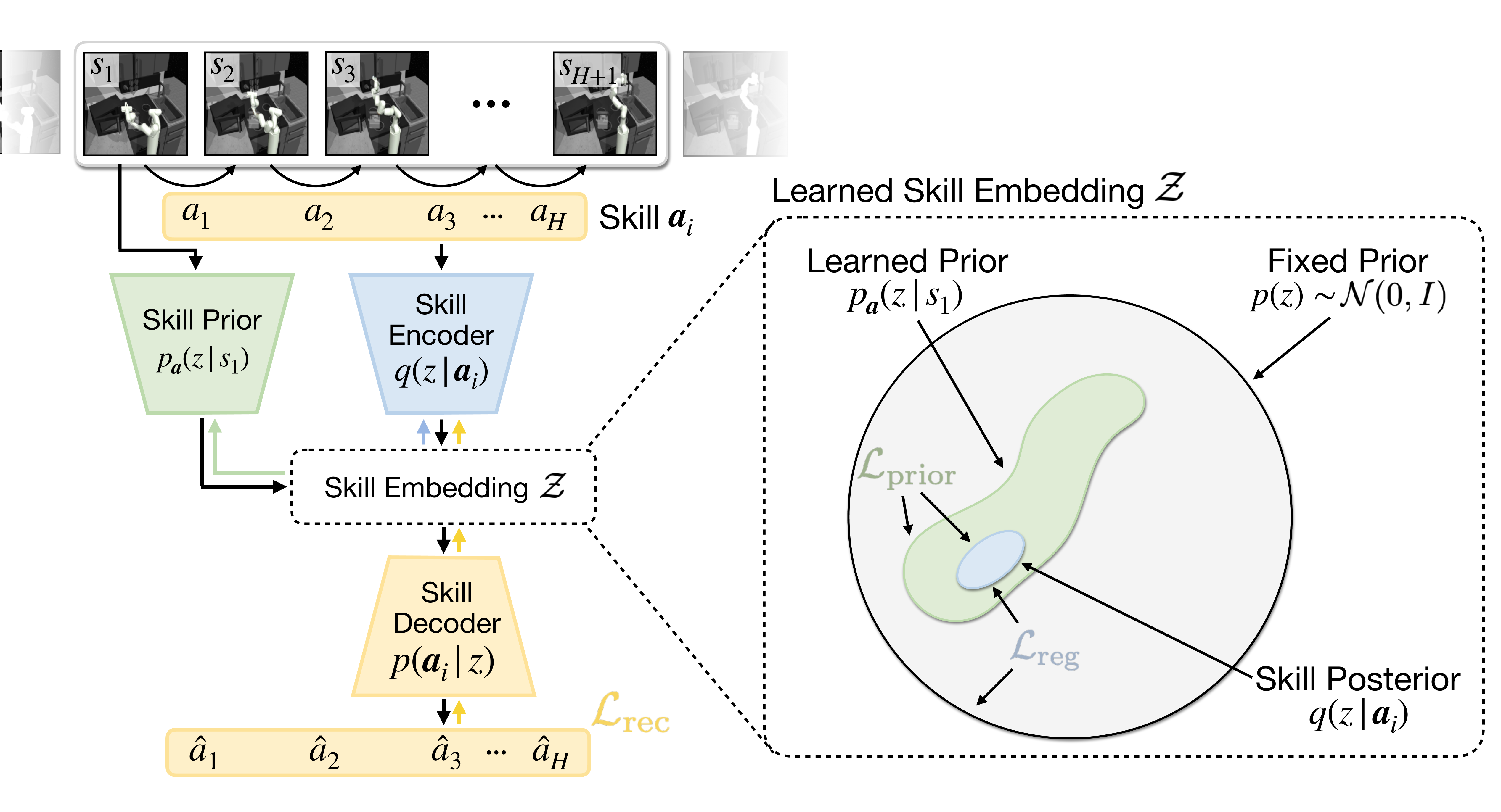}
    \caption{Deep latent variable model for joint learning of skill embedding and skill prior. Given a state-action trajectory from the dataset, the skill encoder maps the action sequence to a posterior distribution $q(z \vert \va_i)$ over latent skill embeddings. The action trajectory gets reconstructed by passing a sample from the posterior through the skill decoder. The skill prior maps the current environment state to a prior distribution $p_{\va}(z \vert s_1)$ over skill embeddings. Colorful arrows indicate the propagation of gradients from \textcolor[HTML]{F4C430}{reconstruction}, \textcolor[HTML]{0F52BA}{regularization} and \textcolor[HTML]{00A86B}{prior training} objectives.
    }
    \label{fig:model}
\end{figure}

We define a skill $\va_i$ as a sequence of actions $\{a_t^i, \dots, a_{t+H-1}^i\}$ with fixed horizon $H$. Using fixed-length skills allows for scalable skill learning and has proven to be effective in prior works~\cite{merel2018neural,merel2019reusable,gupta2019relay,whitney2019dynamics,mandlekar2019iris,fang2019dynamics}. Other work has proposed to learn semantic skills of flexible length~\cite{kipf2018compositional,shankar2019discovering,pertsch2020keyin} %
and our model can be extended to include similar approaches, but we leave this for future work.

To learn a low-dimensional skill embedding space~$\mathcal{Z}$, we train a stochastic latent variable model $p(\va_i \vert z)$ of skills using the offline dataset (see Fig.~\ref{fig:model}). We randomly sample $H$-step trajectories from the training sequences and maximize the following evidence lower bound (ELBO):
\begin{equation}
    \log p(\va_i) \geq \mathbb{E}_q \bigg[\underbrace{\log p(\va_i \vert z)}_{\text{reconstruction}} - \beta \big(\underbrace{\log q(z \vert \va_i) - \log p(z)}_{\text{regularization}}\big) \bigg].
\end{equation}
Here, $\beta$ is a parameter that is commonly used to tune the weight of the regularization term~\cite{higgins2017beta}. We optimize this objective using amortized variational inference with an inference network~$q(z \vert \va_i)$~\cite{kingma2014auto,rezende2014stochastic}. To learn a rich skill embedding space we implement skill encoder~$q(z \vert \va_i)$ and decoder~$p(\va_i \vert z)$ as deep neural networks that output the parameters of the Gaussian posterior and output distributions. The prior $p(z)$ is set to be unit Gaussian~$\mathcal{N}(0, I)$. Once trained, we can sample skills from our model by sampling latent variables $z \sim \mathcal{N}(0, I)$ and passing them through the decoder~$p(\va_i \vert z)$. In Section~\ref{sec:skill_prior_rl} we show how to use this generative model of skills for learning hierarchical RL policies.

To better guide downstream learning, we learn a prior over skills along with the skill embedding model. We therefore introduce another component in our model: the skill prior $p_\va(z \vert \cdot)$. The conditioning of this skill prior can be adjusted to the environment and task at hand, but should be informative about the set of skills that are meaningful to explore in a given situation. Possible choices include the embedding of the last executed skill $z_{t-1}$ or the current environment state $s_t$. In this work we focus on learning a state-conditioned skill prior $p_\va(z \vert s_t)$. Intuitively, the current state should provide a strong prior over which skills are promising to explore and, importantly, which skills should not be explored in the current situation (see Fig.~\ref{fig:overview}).

To train the skill prior we minimize the Kullback-Leibler divergence between the predicted prior and the inferred skill posterior: $\mathbb{E}_{(s, \va_i) \sim \mathcal{D}}  D_{\text{KL}}\big(q(z \vert \va_i), p_\va(z \vert s_t)\big)$. Using the reverse KL divergence $D_\text{KL}(q,p)$ instead of $D_\text{KL}(p,q)$ ensures that the learned prior is mode-covering~\cite{bishop2006pattern}, %
\ie represents \emph{all} observed skills in the current situation. Instead of training the skill prior after training the skill embedding model, we can jointly optimize both models and ensure stable convergence by stopping gradients from the skill prior objective into the skill encoder. We experimented with different parameterizations for the skill prior distribution, in particular multi-modal distributions, such as Gaussian mixture models and normalizing flows~\cite{rezende2015variational,dinh2016density}, but found simple Gaussian skill priors to work equally well in our experiments. For further implementation details, see appendix, Section~\ref{sec:implementation_details}.

\subsection{Skill Prior Regularized Reinforcement Learning}
\label{sec:skill_prior_rl}

\begin{algorithm}[t]
\caption{\spirl: Skill-Prior RL} %
\label{alg:skill_prior_sac}
\begin{algorithmic}[1]
\STATE \textbf{Inputs:} $H$-step reward function $\tilde{r}(s_t, z_t)$, discount $\gamma$, target divergence $\delta$, learning rates $\lambda_{\pi}, \lambda_Q, \lambda_\alpha$, target update rate $\tau$.
\STATE Initialize replay buffer $\mathcal{D}$, high-level policy $\pi_\theta(z_t\vert s_t)$, critic $Q_\phi(s_t, z_t)$, target network $Q_{\bar{\phi}}(s_t, z_t)$
\FOR{each iteration}

\FOR{every $H$ environment steps}
\STATE $z_t \sim \pi(z_t \vert s_t)$ \COMMENT{sample skill from policy}
\STATE $s_{t^\prime} \sim p(s_{t+H} \vert s_t, z_t)$ \COMMENT{execute skill in environment}
\STATE $\mathcal{D} \leftarrow \mathcal{D} \cup \{s_t, z_t, \tilde{r}(s_t, z_t), s_{t^\prime}\}$ \COMMENT{store transition in replay buffer}
\ENDFOR

\FOR{each gradient step}
\STATE $\bar{Q} = \tilde{r}(s_t, z_t) + \gamma \big[ Q_{\bar{\phi}}(s_{t^\prime}, \pi_\theta(z_{t^\prime} \vert s_{t^\prime})) \textcolor{red}{- \alpha D_\text{KL}\big(\pi_\theta(z_{t^\prime} \vert s_{t^\prime}), p_\va(z_{t^\prime} \vert s_{t^\prime})\big)} \big]$ \COMMENT{compute Q-target}
\STATE $\theta \leftarrow \theta - \lambda_\pi \nabla_\theta \big[Q_\phi(s_t, \pi_\theta(z_t \vert s_t)) \textcolor{red}{- \alpha D_\text{KL}(\pi_\theta(z_t \vert s_t), p_\va(z_t \vert s_t))}\big] $ \COMMENT{update policy weights}
\STATE $\phi \leftarrow \phi - \lambda_Q \nabla_\phi \big[ \frac{1}{2}\big(Q_\phi(s_t, z_t) - \bar{Q} \big)^2 \big]$ \COMMENT{update critic weights}
\STATE $\alpha \leftarrow \alpha - \lambda_\alpha \nabla_\alpha \big[ \alpha \cdot (\textcolor{red}{D_\text{KL}(\pi_\theta(z_t \vert s_t), p_\va(z_t \vert s_t)) - \delta}) \big]$ \COMMENT{update alpha}
\STATE $\bar{\phi} \leftarrow \tau \phi + (1 - \tau) \bar{\phi}$ \COMMENT{update target network weights}
\ENDFOR

\ENDFOR
\STATE \textbf{return} trained policy $\pi_\theta(z_t \vert s_t)$
\end{algorithmic}
\end{algorithm}

To use the learned skill embedding for downstream task learning, we employ a hierarchical policy learning scheme by using the skill embedding space as action space of a high-level policy. Concretely, instead of learning a policy over actions $a \in \mathcal{A}$ we learn a policy $\pi_\theta(z \vert s_t)$ that outputs skill embeddings, which we decode into action sequences using the learned skill decoder $\{a_t^i, \dots, a_{t+H-1}^i\} \sim p(\va_i \vert z)$\footnote{We also experimented with models that directly condition the decoder on the current state, but found downstream RL to be less stable (see appendix, Section~\ref{sec:state_cond_decoder})}.
We execute these actions for $H$ steps before sampling the next skill from the high-level policy. This hierarchical structure allows for temporal abstraction, which facilitates long-horizon task learning~\cite{sutton1999between}.

We can cast the problem of learning the high-level policy into a standard MDP by replacing the action space $\mathcal{A}$ with the skill space $\mathcal{Z}$, single-step rewards with $H$-step rewards $\tilde{r} = \sum_{t=1}^H r_t$, and single-step transitions with $H$-step transitions $s_{t+H} \sim p(s_{t+H} \vert s_t, z_t)$. We can then use conventional model-free RL approaches to maximize the return of the high-level policy $\pi_\theta(z \vert s_t)$.

This naive approach struggles when training a policy on a very rich skill space $\mathcal{Z}$ that encodes many different skills. While the nominal dimensionality of the skill space might be small, its continuous nature allows the model to embed an arbitrary number of different behaviors. Therefore, the \emph{effective} dimensionality of the high-level policies' action space scales with the number of embedded skills. When using a large offline dataset $\mathcal{D}$ with diverse behaviors, the number of embedded skills can grow rapidly, leading to a challenging exploration problem when training the high-level policy. For more efficient exploration, we propose to use the learned skill prior to guide the high-level policy. We will next show how the skill prior can be naturally integrated into maximum-entropy RL algorithms.

Maximum entropy RL~\cite{ziebart2010modeling,levine2018reinforcement} %
augments the training objective of the policy with a term that encourages maximization of the policy's entropy along with the return:
\begin{equation}
    J(\theta) = \mathbb{E}_\pi \bigg[\sum_{t=1}^T \gamma^t r(s_t, a_t) + \alpha \mathcal{H}\big(\pi(a_t \vert s_t)\big)\bigg]
\end{equation}

The added entropy term is equivalent to the negated KL divergence between the policy and a \emph{uniform} action prior $U(a_t)$: $\mathcal{H}(\pi(a_t \vert s_t)) = - \mathbb{E}_\pi \big[\log \pi(a_t \vert s_t)\big] \propto -D_{\text{KL}}(\pi(a_t \vert s_t), U(a_t))$ up to a constant. However, in our case we aim to regularize the policy towards a \emph{non-uniform}, learned skill prior to guide exploration in skill space. 
We can therefore replace the entropy term with the negated KL divergence from the learned prior, leading to the following objective for the high-level policy:
\begin{equation}
    J(\theta) = \mathbb{E}_\pi \bigg[\sum_{t=1}^T \tilde{r}(s_t, z_t) - \alpha D_{\text{KL}}\big(\pi(z_t \vert s_t), p_\va(z_t \vert s_t)\big)\bigg]
\end{equation}

We can modify the state-of-the-art maximum-entropy RL algorithms, such as Soft Actor-Critic (SAC~\cite{haarnoja2018sac,haarnoja2018sac_algo_applications}) to optimize this objective. We summarize our \spirl~approach in Algorithm~\ref{alg:skill_prior_sac} with changes to SAC marked in \textcolor{red}{red}. For a detailed derivation of the update rules, see appendix, Section~\ref{sec:action_prior_sac}. Analogous to~\citet{haarnoja2018sac_algo_applications} we can devise an automatic tuning strategy for the regularization weight $\alpha$ by defining a \emph{target divergence} parameter $\delta$ (see Algorithm~\ref{alg:skill_prior_sac}, appendix~\ref{sec:action_prior_sac}).

%% file: sections/experiments.tex
\section{Experiments}
\label{sec:experiments}

Our experiments are designed to answer the following questions: (1) Can we leverage unstructured datasets to accelerate downstream task learning by transferring skills? (2) Can learned skill priors improve exploration during downstream task learning? (3) Are learned skill priors necessary to scale skill transfer to large datasets?

\subsection{Environments \& Comparisons}

\begin{figure}[t]
    \centering
    \includegraphics[width=1\linewidth]{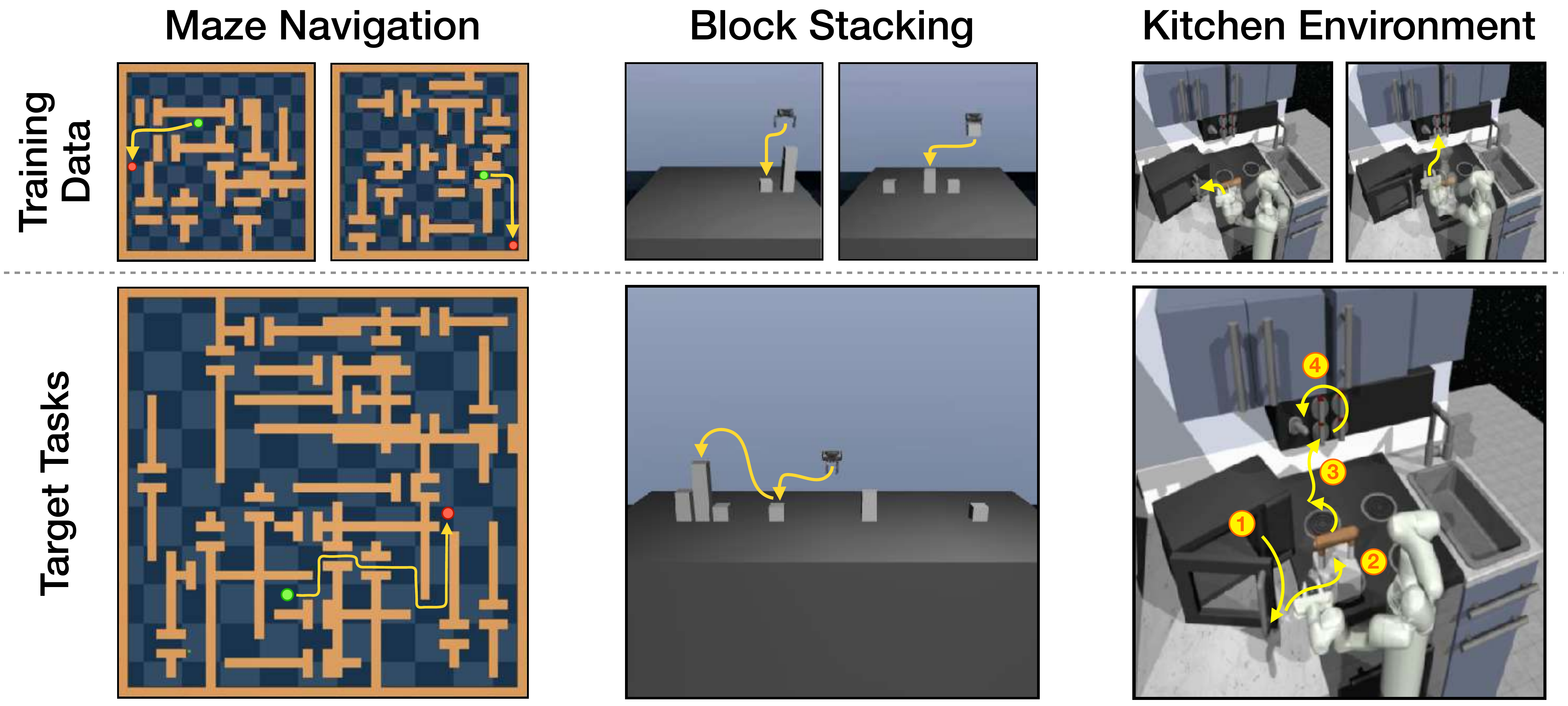}
    \caption{
    For each environment we collect a diverse dataset from a wide range of training tasks (examples on \textbf{top}) and test skill transfer to more complex target tasks (\textbf{bottom}), in which the agent needs to: navigate a maze (\textbf{left}), stack as many blocks as possible (\textbf{middle}) and manipulate a kitchen setup to reach a target configuration (\textbf{right}). All tasks require the execution of complex, long-horizon behaviors and need to be learned from sparse rewards.
    }
    \label{fig:env_overview}
\end{figure}

We evaluate \spirl~on one simulated navigation task and two simulated robotic manipulation tasks (see Fig.~\ref{fig:env_overview}). For each environment, we collect a large and diverse dataset of agent experience that allows to extract a large number of skills. To test our method's ability to transfer to \emph{unseen} downstream tasks, we vary task and environment setup between training data collection and downstream task.

\paragraph{Maze Navigation.} A simulated maze navigation environment based on the D4RL maze environment~\cite{fu2020d4rl}. The task is to navigate a point mass agent through a maze between fixed start and goal locations. We use a planner-based policy to collect \SI{85000}{} goal-reaching trajectories in randomly generated, small maze layouts and test generalization to a goal-reaching task in a randomly generated, larger maze. The state is represented as a RGB top-down view centered around the agent. For downstream learning the agent only receives a sparse reward when in close vicinity to the goal. The agent can transfer skills, such as traversing hallways or passing through narrow doors, but needs to learn to navigate a new maze layout for solving the downstream task.

\paragraph{Block Stacking.} The goal of the agent is to stack as many blocks as possible in an environment with eleven blocks. 
We collect \SI{37000}{} training sequences with a noisy, scripted policy that randomly stacks blocks on top of each other in a smaller environment with only five blocks. The state is represented as a RGB front view centered around the agent and it receives binary rewards for picking up and stacking blocks. The agent can transfer skills like picking up, carrying and stacking blocks, but needs to perform a larger number of consecutive stacks than seen in the training data on a new environment with more blocks.

\paragraph{Kitchen Environment.} A simulated kitchen environment based on~\citet{gupta2019relay}. We use the training data provided in the D4RL benchmark~\cite{fu2020d4rl}, which consists of \SI{400}{} teleoperated sequences in which the 7-DoF robot arm manipulates different parts of the environment (\eg open microwave, switch on stove, slide cabinet door). During downstream learning the agent needs to execute an unseen sequence of multiple subtasks. It receives a sparse, binary reward for each successfully completed manipulation. The agent can transfer a rich set of manipulation skills, but needs to recombine them in new ways to solve the downstream task.

For further details on environment setup, data collection and training, see appendix, Sections~\ref{sec:implementation_details}~and~\ref{sec:env_data_details}.

We compare the downstream task performance of \spirl~to several flat and hierarchical baselines that test the importance of learned skill embeddings and skill prior:
\begin{itemize}
    \item \textbf{Flat Model-Free RL (SAC).} Trains an agent \emph{from scratch} with Soft Actor-Critic (SAC,~\cite{haarnoja2018sac}). %
    This comparison tests the benefit of leveraging prior experience.
    
    \item \textbf{Behavioral Cloning w/ finetuning (BC + SAC).} Trains a supervised behavioral cloning (BC) policy from the offline data and finetunes it on the downstream task using SAC.
    
    \item \textbf{Flat Behavior Prior (Flat Prior).} Learns a single-step action prior on the primitive action space and uses it to regularize downstream learning as described in Section~\ref{sec:skill_prior_rl}, similar to~\cite{siegel2020keep}. This comparison tests the importance of temporal abstraction through learned skills.
    
    \item \textbf{Hierarchical Skill-Space Policy (SSP).} Trains a high-level policy on the skill-embedding space of the model described in Section~\ref{sec:skill_embed_prior_learning} but without skill prior, representative of~\cite{merel2018neural,kipf2018compositional,shankar2019discovering}. This comparison tests the importance of the learned skill prior for downstream task learning.

\end{itemize}

\begin{figure}[t]
    \centering
    \includegraphics[width=1\textwidth]{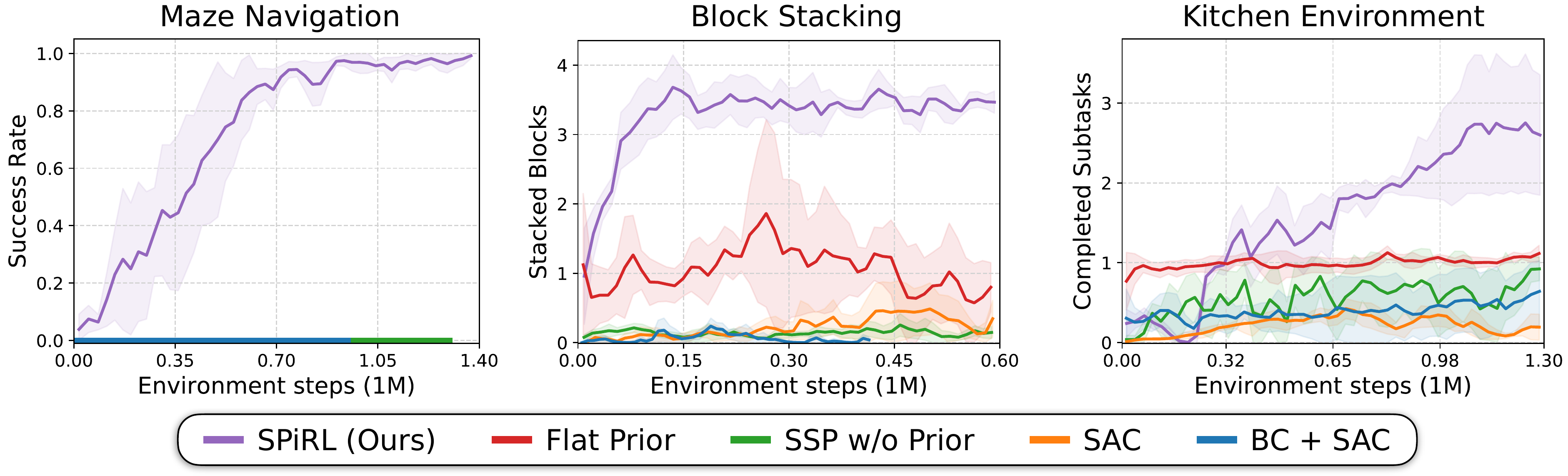}
    \vspace{-10pt}
    \caption{
    Downstream task learning curves for our method and all comparisons. Both, learned skill embedding and skill prior are essential for downstream task performance: single-action priors without temporal abstraction (\textbf{Flat Prior}) and learned skills without skill prior (\textbf{SSP w/o Prior}) fail to converge to good performance. Shaded areas represent standard deviation across three seeds.
    }
    \label{fig:training_curves}
    \vspace{-10pt}
\end{figure}

\subsection{Maze Navigation}

We first evaluate \spirl~on the simulated maze navigation task. This task poses a hard exploration problem since the reward feedback is very sparse: following the D4RL benchmark~\cite{fu2020d4rl} the agent receives a binary reward only when reaching the goal and therefore needs to explore large fractions of the maze without reward feedback. We hypothesize that learned skills and a prior that guides exploration are crucial for successful learning, particularly when external feedback is sparse.

\begin{figure}
    \centering
    \includegraphics[width=1\linewidth]{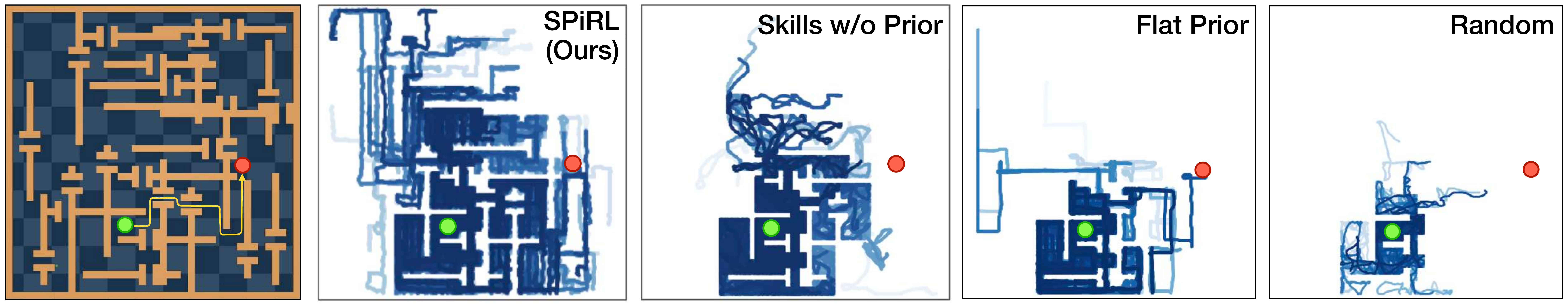}
    \vspace{-10pt}
    \caption{Exploration behavior of our method vs. alternative transfer approaches on the downstream maze task vs. random action sampling. Through learned skill embeddings and skill priors our method can explore the environment more widely. We visualize positions of the agent during 1M steps of exploration rollouts in blue and mark episode start and goal positions in green and red respectively.
    }
    \vspace{-10pt}
    \label{fig:exploration}
\end{figure}

In Fig.~\ref{fig:training_curves} (left) we show that only \spirl~is able to successfully learn a goal-reaching policy for the maze task; none of the baseline policies reaches the goal during training. To better understand this result, we compare the exploration behaviors of our approach and the baselines in Fig.~\ref{fig:exploration}: we collect rollouts by sampling from our skill prior and the single-step action prior and record the agent's position in the maze. To visualize the exploration behavior of skill-space policies without learned priors ("Skills w/o Prior") we sample skills uniformly from the skill space.

Fig.~\ref{fig:exploration} shows that only \spirl~is able to explore large parts of the maze, since targeted sampling of skills from the prior guides the agent to navigate through doorways and traverse hallways. Random exploration in skill space, in contrast, does not lead to good exploration behavior since %
the agent often samples skills that lead to collisions. The comparison to single-step action priors ("Flat Prior") shows that temporal abstraction is beneficial for coherent exploration. %

Finally, we show reuse of a single skill prior for a variety of downstream goals in appendix, Section~\ref{sec:prior_reuse}.

\subsection{Robotic Manipulation}
\label{sec:experiments_blocks_kitchen}

Next, we investigate the ability of \spirl~to scale to complex, robotic manipulation tasks in the block stacking problem and in the kitchen environment. For both environments we find that using learned skill embeddings together with the extracted skill prior is essential to solve the task (see Fig.~\ref{fig:training_curves}, middle and right; appendix Fig.~\ref{fig:execution_traces} for qualitative policy rollouts). In contrast, using non-hierarchical action priors ("Flat Prior") leads to performance similar to behavioral cloning of the training dataset, but fails to solve longer-horizon tasks. This shows the benefit of temporal abstraction through skills. The approach leveraging the learned skill space without guidance from the skill prior ("SSP w/o Prior") only rarely stacks blocks or successfully manipulates objects in the kitchen environment. Due to the large number of extracted skills from the rich training datasets, random exploration in skill space does not lead to efficient downstream learning. Instead, performance is comparable or worse than learning from scratch without skill transfer. This underlines the importance of learned skill priors for scaling skill transfer to large datasets. Similar to prior work~\cite{gupta2019relay}, we find that a policy initialized through behavioral cloning is not amenable to efficient finetuning on complex, long-horizon tasks.

\subsection{Ablation Studies}

\begin{wrapfigure}{R}{0.6\textwidth}
\vspace{-15pt}
    \begin{minipage}{0.29\textwidth}
        \centering
        \includegraphics[width=1\textwidth]{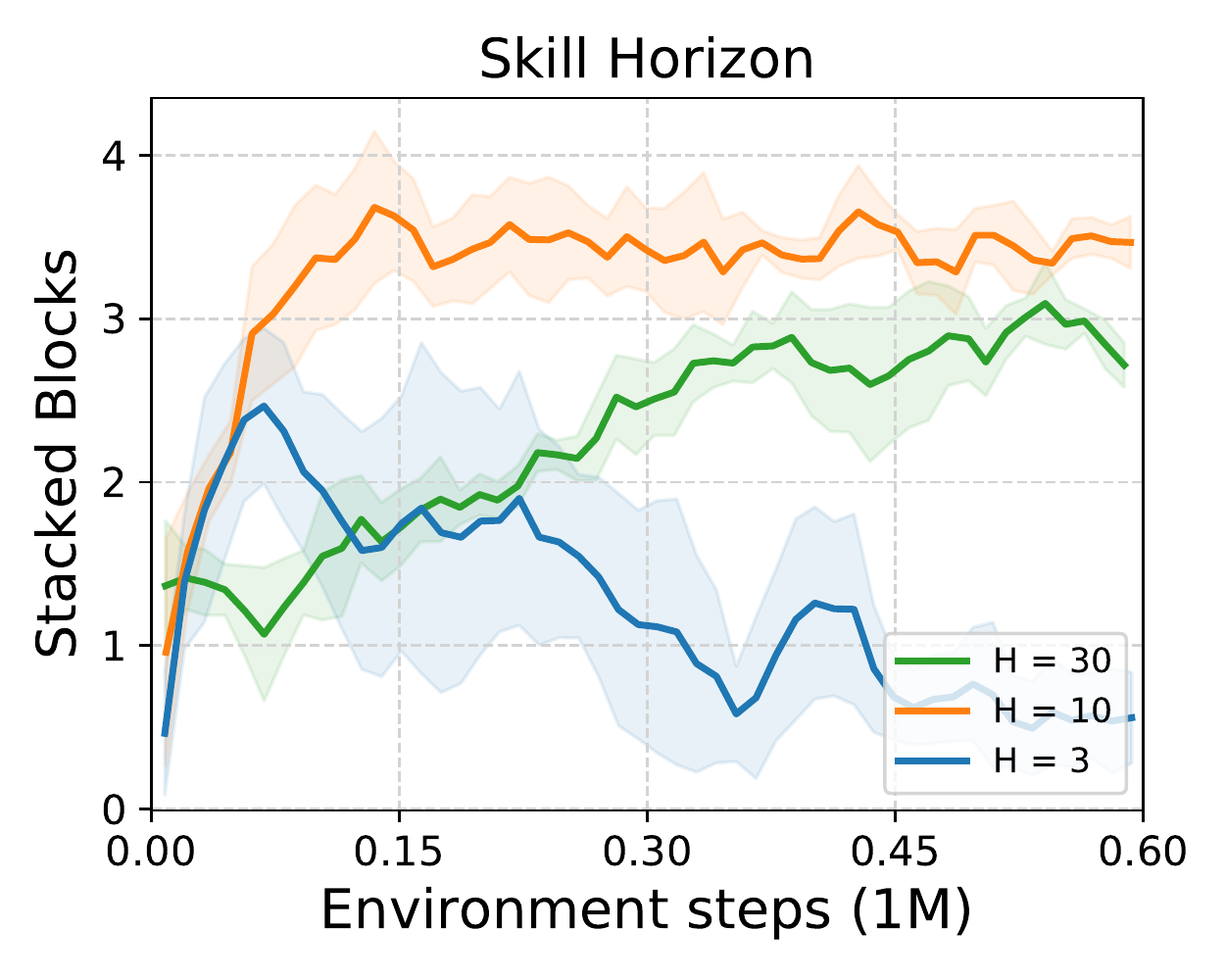}
    \end{minipage}
    \hfill
    \begin{minipage}{0.29\textwidth}
        \centering
        \includegraphics[width=1\textwidth]{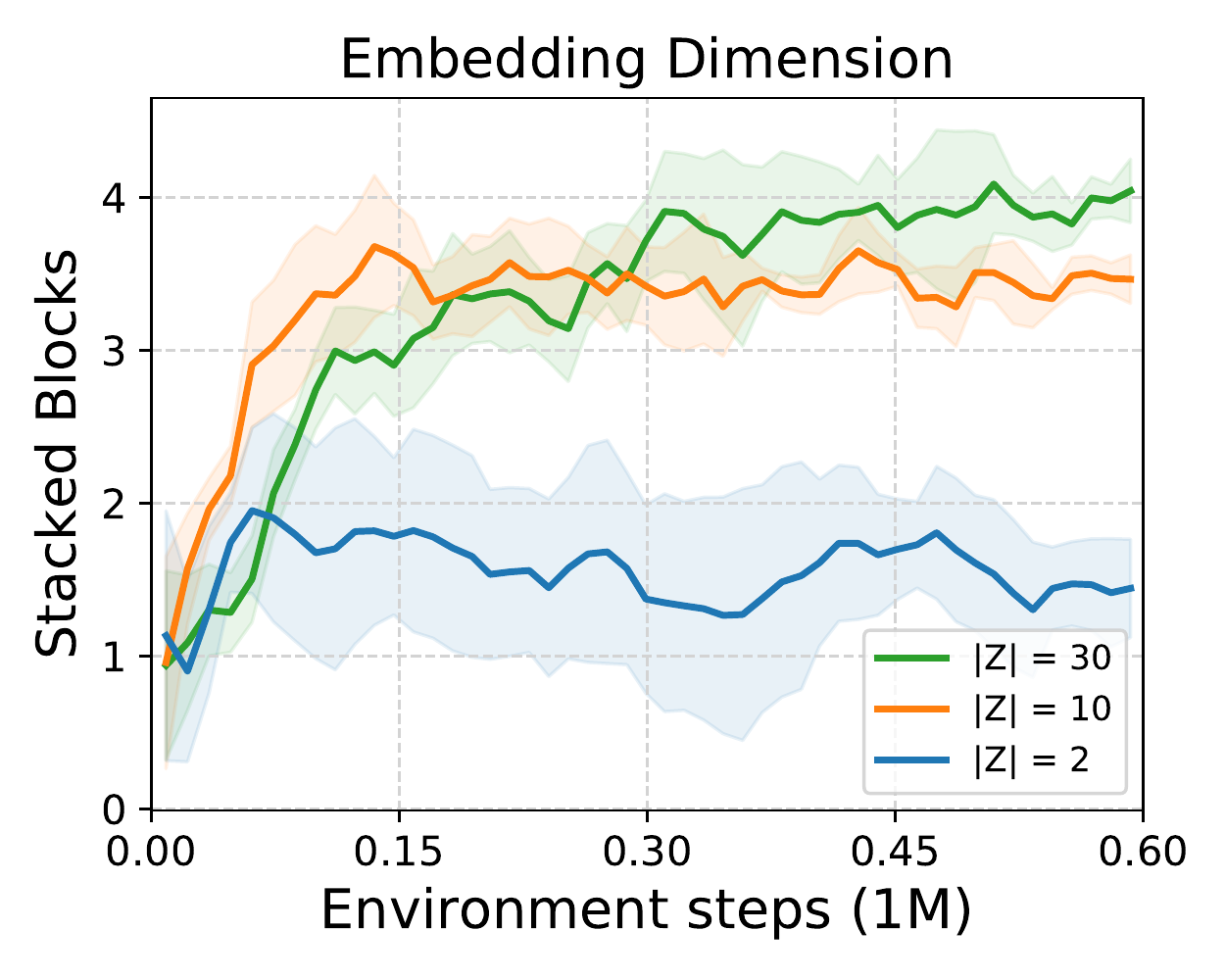}
    \end{minipage}
    \caption{Ablation analysis of skill horizon and skill space dimensionality on block stacking task.
    }
    \label{fig:ablations}
    \vspace{-10pt}
\end{wrapfigure}

We analyze the influence of skill horizon $H$ %
and dimensionality of the learned skill space $\vert \mathcal{Z} \vert$ on downstream performance in Fig.~\ref{fig:ablations}. We see that too short skill horizons do not afford sufficient temporal abstraction. %
Conversely, too long horizons make the skill exploration problem harder, since a larger number of possible skills gets embedded in the skill space. Therefore, the policy converges slower. %

We find that the dimensionality of the learned skill embedding space needs to be large enough to represent a sufficient diversity of skills. Beyond that, $\vert \mathcal{Z} \vert$ does not have a major influence on the downstream performance. We attribute this to the usage of the learned skill prior: even though the nominal dimensionality of the high-level policy's action space increases, its effective dimensionality remains unchanged since the skill prior focuses exploration on the relevant parts of the skill space.

We further test the importance of prior initialization and regularization, as well as training priors from sub-optimal data in appendix, Sections~\ref{sec:ablate_prior_reg}~-~\ref{sec:subopt_data}.

%% file: sections/conclusion.tex
\section{Conclusion}
\label{sec:conclusion}
We present \spirl, an approach for leveraging large, unstructured datasets to accelerate downstream learning of unseen tasks. We propose a deep latent variable model that jointly learns an embedding space of skills and a prior over these skills from offline data. We then extend maximum-entropy RL algorithms to incorporate both skill embedding and skill prior for efficient downstream learning. Finally, we evaluate \spirl~on challenging simulated navigation and robotic manipulation tasks and show that both, skill embedding and skill prior are essential for effective transfer from rich datasets.

Future work can combine learned skill priors with methods for extracting \emph{semantic} skills of flexible length from unstructured data~\cite{shankar2019discovering,pertsch2020keyin}. %
Further, skill priors are important in safety-critical applications, like autonomous driving, where random exploration is dangerous. Skill priors learned e.g. from human demonstration, can guide exploration to skills that do not endanger the learner or other agents.

%% file: sections/appendix.tex
\section{Action-prior Regularized Soft Actor-Critic}
\label{sec:action_prior_sac}

The original derivation of the SAC algorithm assumes a uniform prior over actions. We extend the formulation to the case with a non-uniform action prior $p(a \vert \cdot)$, where the dot indicates that the prior can be non-conditional or conditioned on \eg the current state or the previous action. Our derivation closely follows \citet{haarnoja2018sac} and \citet{levine2018reinforcement} with the key difference that we replace the entropy maximization in the reward function with a term that penalizes divergence from the action prior. We derive the formulation for single-step action priors below, and the extension to \emph{skill priors} is straightforward by replacing actions $a_t$ with skill embeddings $z_t$.

We adopt the probabilistic graphical model~(PGM) described in \cite{levine2018reinforcement}, which includes \emph{optimality variables} $\mathcal{O}_{1:T}$, whose distribution is defined as $p(\mathcal{O}_t \vert s_t, a_t) = \exp{\big(r(s_t, a_t)\big)}$ where $r(s_t, a_t)$ is the reward. We treat $\mathcal{O}_{1:T} = 1$ as evidence in our PGM and obtain the following conditional trajectory distribution:
\begin{align*}
    p(\tau \vert \mathcal{O}_{1:T}) &= p(s_1) \prod_{t=1}^T p(\mathcal{O}_t \vert s_t, a_t) p(s_{t+1} \vert s_t, a_t) p(a_t\vert \cdot) \\
                                    &= \bigg[ p(s_1) \prod_{t=1}^T p(s_{t+1} \vert s_t, a_t) p(a_t\vert \cdot)\bigg] \cdot \exp{\sum_{t=1}^T r(s_t, a_t)}
\end{align*}
Crucially, in contrast to \cite{levine2018reinforcement} we did not omit the action prior $p(a_t\vert \cdot)$ since we assume it to be generally not uniform.

Our goal is to derive an objective for learning a policy that induces such a trajectory distribution. Following \cite{levine2018reinforcement} we will cast this problem within the framework of structured variational inference and derive an expression for an evidence lower bound (ELBO).

We define a variational distribution $q(a_t \vert s_t)$ that represents our policy. It induces a trajectory distribution $q(\tau) = p(s_1) \prod_{t = 1}^T p(s_{t+1} \vert s_t, a_t) q(a_t \vert s_t)$. We can derive the ELBO as:
\begin{align*}
    \log p(\mathcal{O}_{1:T}) &\geq -D_{\text{KL}}\big[ q(\tau) \;\vert\vert\; p(\tau \vert \mathcal{O}_{1:T})\big] \\
                              &\geq \mathbb{E}_{\tau \sim q(\tau)} \Bigg[\log p(s_1) + \sum_{t=1}^T \big[\log p(s_{t+1}\vert s_t, a_t) \log p(a_t \vert \cdot)\big] + \sum_{t=1}^T r(s_t, a_t) \\
                              & \quad\quad\quad\quad - \log p(s_1) - \sum_{t=1}^T \big[\log p(s_{t+1}\vert s_t, a_t) \log q(a_t \vert s_t) \big] \Bigg] \\
                              &\geq \mathbb{E}_{\tau \sim q(\tau)} \Bigg[\sum_{t=1}^T r(s_t, a_t) + \log p(a_t \vert \cdot) - \log q(a_t \vert s_t)\Bigg] \\
                              &\geq \mathbb{E}_{\tau \sim q(\tau)} \Bigg[\sum_{t=1}^T r(s_t, a_t) - D_{\text{KL}}\big[q(a_t \vert s_t) \;\vert\vert\; p(a_t \vert \cdot)\big]\Bigg]
\end{align*}
Note that in the case of a uniform action prior the KL divergence is equivalent to the negative entropy $- \mathcal{H}(q(a_t \vert s_t))$. Substituting the KL divergence with the entropy recovers the ELBO derived in \cite{levine2018reinforcement}.

To maximize this ELBO with respect to the policy $q(a_t \vert s_t)$, \cite{levine2018reinforcement} propose to use an inference procedure based on a message passing algorithm. 
Following this derivation for the "messages" $V(s_t)$ and $Q(s_t, a_t)$ (\citet{levine2018reinforcement}, Section~4.2), but substituting policy entropy $-\log q(a_t \vert s_t)$ with prior divergence $D_{\text{KL}}\big[q(a_t \vert s_t) \;\vert\vert\; p(a_t \vert \cdot)\big]$, the \textbf{modified Bellman backup operator} can be derived as:
\begin{align*}
    &\mathcal{T}^\pi Q(s_t, a_t) = r(s_t, a_t) + \gamma \mathbb{E}_{s_{t+1} \sim p} \big[ V(s_{t+1}) \big] \\
           \text{where }& V(s_t) = \mathbb{E}_{a_t \sim \pi} \big[ Q(s_t, a_t) - D_{\text{KL}}\big[\pi(a_t \vert s_t) \;\vert\vert\; p(a_t \vert \cdot) \big] \big]
\end{align*}
To show convergence of this operator to the optimal Q function we follow the proof of \cite{haarnoja2018sac} in appendix B1 and introduce a divergence-augmented reward:
\begin{equation*}
    r_\pi(s_t, a_t) = r(s_t, a_t) - \mathbb{E}_{s_{t+1} \sim p} \bigg[ D_{\text{KL}}\big[\pi(a_{t+1} \vert s_{t+1}) \;\vert\vert\; p(a_{t+1} \vert \cdot) \big] \bigg].
\end{equation*}
Then we can recover the original Bellman update: 
\begin{equation*}
    Q(s_t, a_t) \leftarrow r_\pi(s_t, a_t) + \gamma \mathbb{E}_{s_{t+1} \sim p, a_{t+1} \sim \pi}\big[ Q(s_{t+1}, a_{t+1}) \big],
\end{equation*}
for which the known convergence guarantees hold \cite{sutton2018reinforcement}. 

The modifications to the messages $Q(s_t, a_t)$ and $V(s_t)$ directly lead to the following \textbf{modified policy improvement operator}:
\begin{equation*}
    \arg\min_\theta \mathbb{E}_{s_t \sim \mathcal{D}, a_t \sim \pi} \bigg[ D_{\text{KL}}\big[\pi(a_{t} \vert s_{t}) \;\vert\vert\; p(a_{t} \vert \cdot) \big] - Q(s_t, a_t) \bigg]
\end{equation*}

Finally, the practical implementations of SAC introduce a temperature parameter $\alpha$ that trades off between the reward and the entropy term in the original formulation and the reward and divergence term in our formulation. \citet{haarnoja2018sac_algo_applications} propose an algorithm to automatically adjust $\alpha$ by formulating policy learning as a constrained optimization problem. In our formulation we derive a similar update mechanism for $\alpha$. We start by formulating the following constrained optimization problem:
\begin{equation*}
    \max_{x_{1:T}} \mathbb{E}_{p_\pi} \bigg[\sum_{t=1}^T r(s_t, a_t)\bigg] \quad \text{s.t. } D_{\text{KL}}\big[\pi(a_{t} \vert s_{t}) \;\vert\vert\; p(a_{t} \vert \cdot) \big] \leq \delta \quad \forall t
\end{equation*}
Here $\delta$ is a target divergence between policy and action prior similar to the target entropy $\bar{H}$ is the original SAC formulation. We can formulate the dual problem by introducing the temperature $\alpha$:
\begin{equation*}
    \min_{\alpha > 0} \max_{\pi} \sum_{t=1}^T r(s_t, a_t) + \alpha \big( \delta - D_{\text{KL}}\big[\pi(a_{t} \vert s_{t}) \;\vert\vert\; p(a_{t} \vert \cdot) \big] \big)
\end{equation*}
This leads to the \textbf{modified update objective for $\alpha$}:
\begin{equation*}
    \arg\min_{\alpha > 0} \mathbb{E}_{a_t \sim \pi} \big[ \alpha \delta - \alpha D_{\text{KL}}\big[\pi(a_{t} \vert s_{t}) \;\vert\vert\; p(a_{t} \vert \cdot) \big] \big]
\end{equation*}
We combine the modified objectives for $Q$-value function, policy and temperature $\alpha$ in the skill-prior regularized SAC algorithm, summarized in Algorithm~\ref{alg:skill_prior_sac}.

\section{Implementation Details}
\label{sec:implementation_details}

\subsection{Model Architecture and Training Objective}

We instantiate the skill embedding model described in Section~\ref{sec:skill_embed_prior_learning} with deep neural networks. The skill encoder is implemented as a one-layer LSTM %
with 128 hidden units. After processing the full input action sequence, it outputs the parameters $(\mu_z, \sigma_z)$ of the Gaussian posterior distribution in the 10-dimensional skill embedding space $\mathcal{Z}$. The skill decoder mirrors the encoder's architecture and is unrolled for $H$ steps to produce the $H$ reconstructed actions. The sampled skill embedding $z$ is passed as input in every step.

The skill prior is implemented as a 6-layer fully-connected network with 128 hidden units per layer. It parametrizes the Gaussian skill prior distribution $\mathcal{N}(\mu_p, \sigma_p)$. For image-based state inputs in maze and block stacking environment, we first pass the state through a convolutional encoder network with three layers, a kernel size of three and $(8, 16, 32)$ channels respectively. The resulting feature map is flattened to form the input to the skill prior network. 

We use leaky-ReLU activations and batch normalization throughout our architecture. We optimize our model using the RAdam optimizer %
with parameters with $\beta_1=0.9$ and $\beta_2=0.999$, batch size 16 and learning rate \SI{1e-3}{}. Training on a single high-end NVIDIA GPU takes approximately 8 hours. Assuming a unit-variance Gaussian output distribution our full training objective is:
\begin{equation}
    \mathcal{L} = 
    \sum_{i=1}^H \underbrace{(a_i - \hat{a}_i)^2}_{\text{reconstruction}} 
    - \beta \underbrace{D_{\text{KL}}\big(\mathcal{N}(\mu_z, \sigma_z) \vert\vert \mathcal{N}(0, I)\big)}_{\text{regularization}} 
    + \underbrace{D_{\text{KL}}\big(\mathcal{N}(\lfloor\mu_z\rfloor, \lfloor\sigma_z\rfloor) \vert\vert \mathcal{N}(\mu_p, \sigma_p)\big)}_{\text{prior training}}.
\end{equation}
Here $\lfloor \cdot \rfloor$ indicates that gradients flowing through these variables are stopped. For Gaussian distributions the KL divergence can be analytically computed. For non-Gaussian prior parametrizations (e.g. with Gaussian mixture model or normalizing flow priors) we found that sampling-based estimates also suffice to achieve reliable, albeit slightly slower convergence. We tune the weighting parameter~$\beta$ separately for each environment and use $\beta$ = \SI{1e-2}{} for maze and block stacking and $\beta$ = \SI{5e-4}{} for the kitchen environment.

\subsection{Reinforcement Learning Setup}

The architecture of policy and critic mirror the one of the skill prior network. The policy outputs the parameters of a Gaussian action distribution while the critic outputs a single $Q$-value estimate. Empirically, we found it important to initialize the weights of the policy with the pre-trained skill prior weights in addition to regularizing towards the prior (see Section~\ref{sec:ablate_prior_init}). 

We use the hyperparameters of the standard SAC implementation~\cite{haarnoja2018sac} with batch size 256, replay buffer capacity of \SI{1e6}{} and discount factor $\gamma = 0.99$. We collect \SI{5e3}{} warmup rollout steps to initialize the replay buffer before training. We use Adam optimizer %
with $\beta_1=0.9$, $\beta_2=0.999$ and learning rate \SI{3e-4}{} for updating policy, critic and temperature $\alpha$. Analogous to SAC, we train two separate critic networks and compute the $Q$-value as the minimum over both estimates to stabilize training. The corresponding target networks get updated at a rate of $\tau$ = \SI{5e-3}{}. The policies' action range is limited in the range $[-2\dots 2]$ by a $\tanh$ "squashing function" (see~\citet{haarnoja2018sac}, appendix C).

We tune the target divergence $\delta$ separately for each environment and use $\delta$ = 1 for the maze navigation task and $\delta$ = 5 for both robot manipulation tasks.

\section{Environments and Data Collection}
\label{sec:env_data_details}

\begin{wrapfigure}{R}{0.5\textwidth}
    \centering
    \vspace{-15pt}
    \includegraphics[width=1\linewidth]{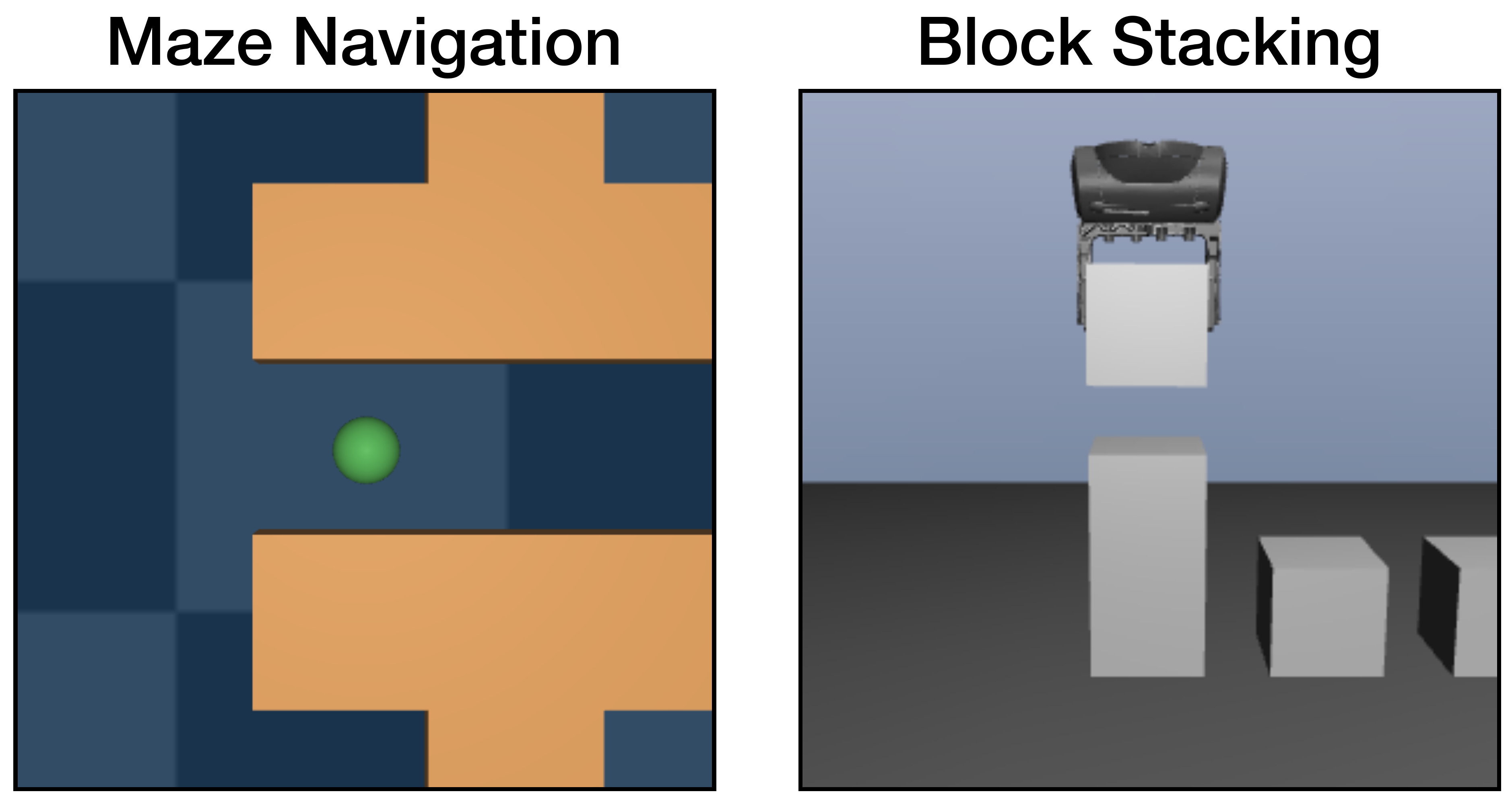}
    \vspace{-15pt}
    \caption{Image-based state representation for maze (\textbf{left}) and block stacking (\textbf{right}) environment (downsampled to $32 \times 32$~px for policy).
    }
    \vspace{-20pt}
    \label{fig:observation_vis}
\end{wrapfigure}

\paragraph{Maze Navigation.} The maze navigation environment is based on the maze environment in the D4RL benchmark~\cite{fu2020d4rl}. Instead of using a single, fixed layout, we generate random layouts for training data collection by placing walls with doorways in randomly sampled positions. For each collected training sequence we sample a new maze layout and randomly sample start and goal position for the agent. Following~\citet{fu2020d4rl}, we collect goal-reaching examples through a combination of high-level planner with access to a map of the maze and a low-level controller that follows the plan. 

For the downstream task we randomly sample a maze that is four times larger than the training data layouts. We keep maze layout, as well as start and goal location for the agent fixed throughout downstream learning. The policy outputs (x,y)-velocities for the agent. The state is represented as a local top-down view around the agent (see Fig.~\ref{fig:observation_vis}). To represent the agent's velocity, we stack two consecutive $32\times 32$px observations as input to the policy. The agent receives a per-timestep binary reward when the distance to the goal is below a threshold.

\paragraph{Block Stacking.} The block stacking environment is simulated using the Mujoco %
physics engine. For data collection, we initialize the five blocks in the environment to be randomly stacked on top of each other or placed at random locations in between. We use a hand-coded data collection policy to generate trajectories with up to three consecutive stacking manipulations. The location of blocks and the movement of the agent are limited to a 2D plane and a barrier prevents the agent from leaving the table. To increase the support of the collected trajectories we add noise to the hard-coded policy by placing pseudo-random subgoals in between and within stacking sequences.

The downstream task of the agent is to stack as many blocks as possible in a larger version of the environment with 11 blocks. The environment state is represented through a front view of the agent (see Fig.~\ref{fig:observation_vis}). The policies' input is a stack of two $32\times 32$px images and it outputs (x,z)-displacements for the robot as well as a continuous action in range $[0\dots 1]$ that represents the opening degree of the gripper. The agent receives per-timestep binary rewards for lifting a block from the ground and moving it on top of another block. It further receives a reward proportional to the height of the highest stacked tower.

\paragraph{Kitchen environment.} We use the kitchen environment from the D4RL benchmark~\cite{fu2020d4rl} which was originally published by~\citet{gupta2019relay}. For training we use the data provided in D4RL (dataset version "mixed"). It consists of trajectories collected via human tele-operation that each perform four consecutive manipulations of objects in the environment. There are seven manipulatable objects in the environment. The downstream task of the agent consists of performing an \emph{unseen} sequence of four manipulations - while the individual manipulations have been observed in the training data, the agent needs to learn to recombine these skills in a new way to solve the task. The state is a 30-dimensional vector representing the agent's joint velocities as well as poses of the manipulatable objects. The agent outputs 7-dimensional joint velocities for robot control as well as a 2-dimensional continuous gripper opening/closing action. It receives a one-time reward whenever fulfilling one of the subtasks.

\begin{figure}
    \centering
    \includegraphics[width=1\linewidth]{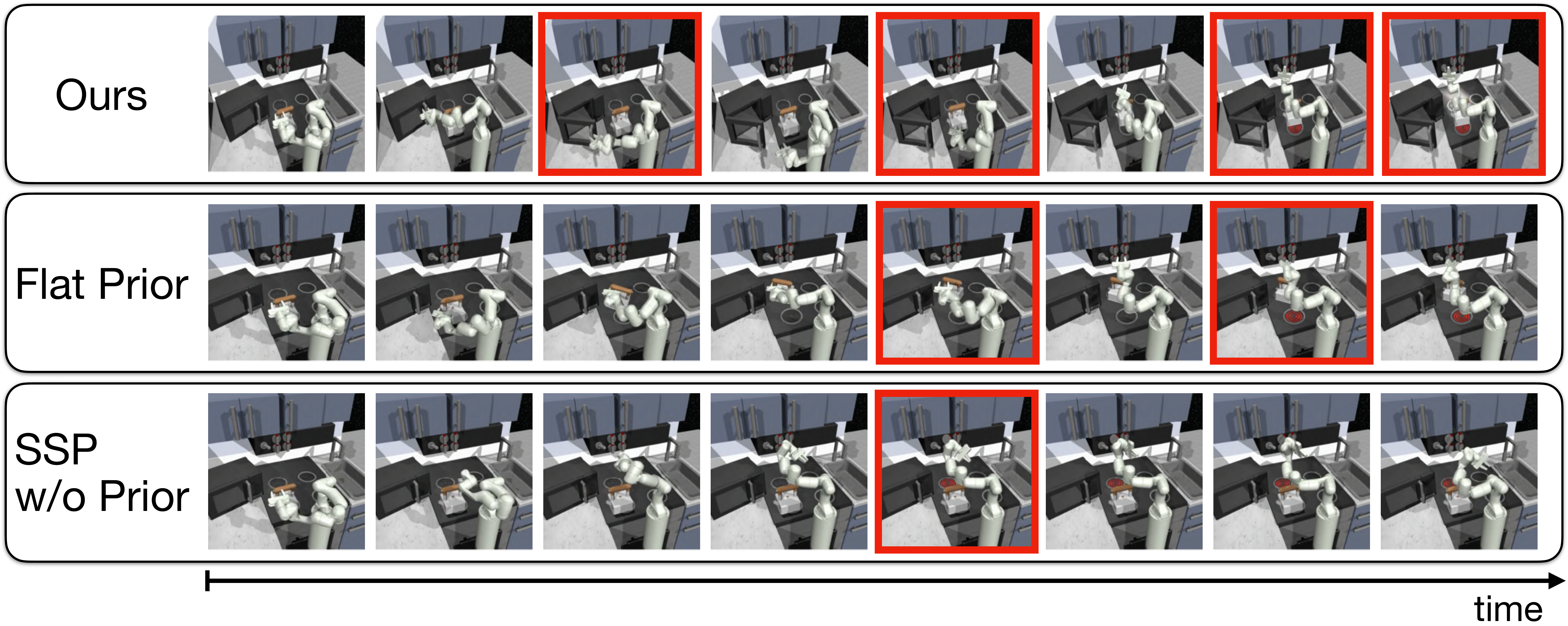}
    \caption{
    Comparison of policy execution traces on the kitchen environment. Following \citet{fu2020d4rl}, the agent's task is to (1)~open the microwave, (2)~move the kettle backwards, (3)~turn on the burner and (4)~switch on the light. Red frames mark the completion of subtasks. Our skill-prior guided agent~(\textbf{top}) is able to complete all four subtasks. In contrast, the agent using a flat single-action prior~(\textbf{middle}) only learns to solve two subtasks, but lacks temporal abstraction and hence fails to solve the complete long-horizon task. The skill-space policy without prior guidance~(\textbf{bottom}) cannot efficiently explore the skill space and gets stuck in a local optimum in which it solves only a single subtask. Best viewed electronically and zoomed in. For videos, see: \href{https://clvrai.com/spirl}{\texttt{clvrai.com/spirl}}.
    }
    \label{fig:execution_traces}
\end{figure}

\section{State-Conditioned Skill Decoder}
\label{sec:state_cond_decoder}

\begin{figure}
    \centering
    \includegraphics[width=1\linewidth]{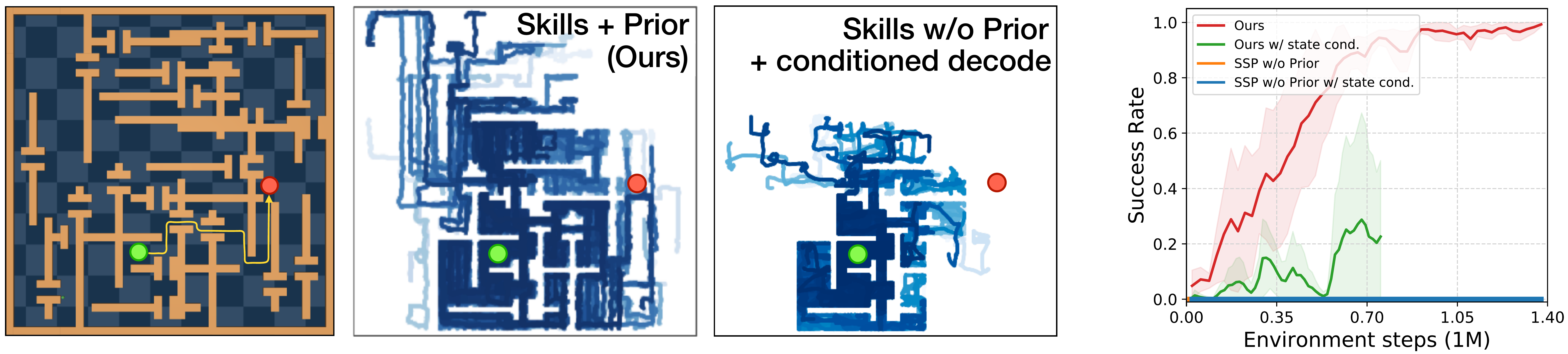}
    \caption{
    Results for state-conditioned skill decoder network. \textbf{left}: Exploration visualization as in Fig.~\ref{fig:exploration}. Even with state-conditioned skill decoder, exploration without skill prior is not able to explore a large fraction of the maze. In contrast, skills sampled from the learned skill prior lead to wide-ranging exploration when using the state-conditioned decoder. \textbf{right}: Downstream learning performance of our approach and skill-space policy w/o learned skill prior: w/ vs. w/o state-conditioning for skill decoder. Only guidance through the learned skill prior enables learning success. State-conditioned skill-decoder can make the downstream learning problem more challenging, leading to lower performance ("ours" vs. "ours w/ state cond.").
    }
    \label{fig:state_cond_maze_results}
\end{figure}

Following prior works~\cite{merel2018neural,kipf2018compositional}, we experimented with conditioning the skill decoder on the current environment state $s_1$. Specifically, the current environment state is passed through a neural network that predicts the initial hidden state of the decoding LSTM. We found that conditioning the skill decoder on the state did not improve downstream learning and can even lead to worse performance. In particular, it does not improve the exploration behavior of the skill-space policy \emph{without} learned prior (see Fig.~\ref{fig:state_cond_maze_results}, left); a learned skill prior is still necessary for efficient exploration on the downstream task. 

Additionally, we found that conditioning the skill decoder on the state can reduce downstream learning performance (see Fig.~\ref{fig:state_cond_maze_results}, right). We hypothesize that state-conditioning can make the learning problem for the high-level policy more challenging: due to the state-conditioning the same high-level action $z$ can result in different decoded action sequences depending on the current state, making the high-level policies' action space dynamics more complex. As a result, downstream learning can be less stable.

\section{Prior Regularization Ablation}
\label{sec:ablate_prior_reg}

\begin{figure}
\centering
\begin{minipage}{.32\textwidth}
  \centering
  \includegraphics[width=\linewidth]{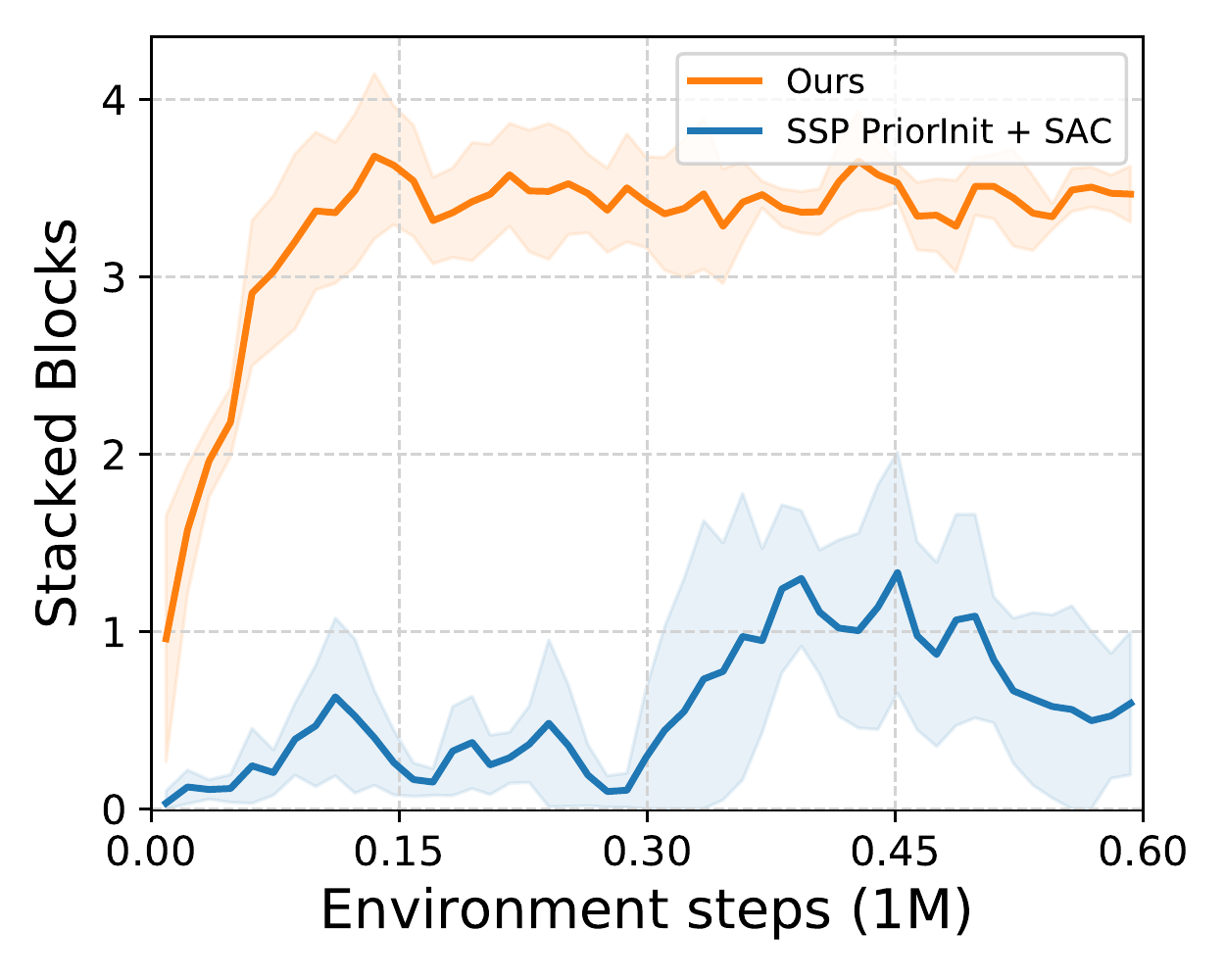}
  \caption{Ablation of prior regularization during downstream RL training. Initializing the high-level policy with the learned prior but finetuning with conventional SAC is not sufficient to learn the task well.
    }
  \label{fig:prior_regularization_ablation}
\end{minipage}%
\hfill
\begin{minipage}{.65\textwidth}
\vspace{-20pt}
  \centering
  \includegraphics[width=\linewidth]{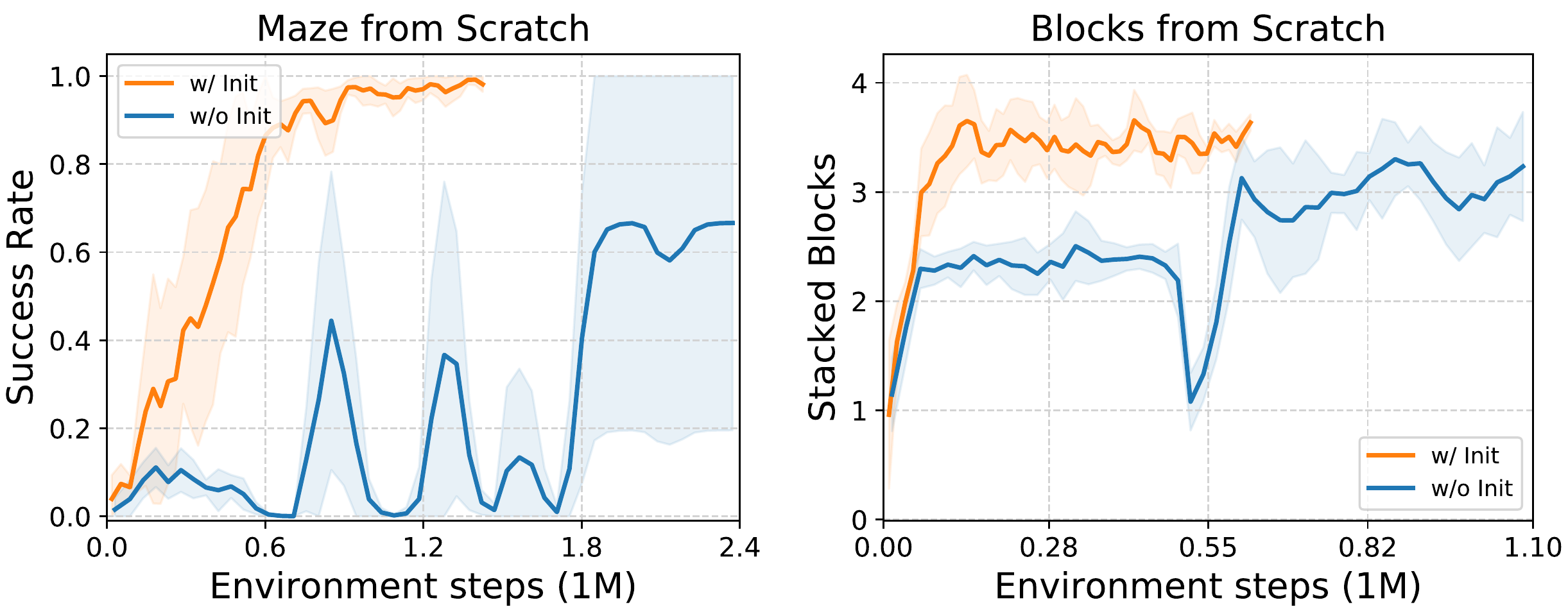}
  \caption{Ablation of prior initialization. Initializing the downstream task policy with the prior network improves training stability and convergence speed. However, the "w/o Init" runs demonstrate that the tasks can also be learned with prior regularization only.}
  \label{fig:from_scratch}
\end{minipage}
\end{figure}

We ablate the influence of the prior regularization during downstream learning as described in Section~\ref{sec:skill_prior_rl}. Specifically, we compare to an approach that initializes the high-level policy with the learned skill prior, but then uses conventional SAC~\cite{haarnoja2018sac} (with uniform skill prior) to finetune on the downstream task. Fig.~\ref{fig:prior_regularization_ablation} shows that the prior regularization during downstream learning is essential for good performance: conventional maximum-entropy SAC ("SSP PriorInit + SAC") quickly leads the prior-initialized policy to deviate from the learned skill prior by encouraging it to maximize the entropy of the distribution over skills, slowing the learning substantially.

\section{Prior Initialization Ablation}
\label{sec:ablate_prior_init}

For the RL experiments in Section~\ref{sec:experiments} we initialize the weights of the high-level policy with the learned skill prior network.
Here, we compare the performance of this approach to an ablation that does not perform the initialization. We find that prior initialization improves convergence speed and stability of training (see Fig.~\ref{fig:from_scratch}).

We identify two major challenges that make training policies "from scratch", without initialization, challenging: (1)~sampling-based divergence estimates between the randomly initialized policy and the prior distribution can be inaccurate in early phases of training, (2)~learning gets stuck in local optima where the divergence between policy and prior is minimized on a small subset of the state space, but the policy does not explore beyond this region.

Since both our learned prior and the high-level policy are parametrized with Gaussian output distributions, we can analytically compute the KL divergence to get a more stable estimate and alleviate problem~(1). To address problem~(2) when training from scratch, we encourage exploration by sampling a fraction $\omega$ of the rollout steps during experience collection directly from the prior instead of the policy. For the "w/o Init" experiments in Fig.~\ref{fig:from_scratch} we set $\omega = 1.0$ for the first \SI{500}{k} steps (i.e. always sample from the prior) and then anneal it to 0 (i.e. always use the policy for rollout collection) over the next \SI{500}{k} steps. Note that policies trained with prior initialization do not require these additions and still converge faster.

\section{Training with Sub-Optimal Data}
\label{sec:subopt_data}

\begin{wraptable}{R}{0.45\linewidth}
\centering
\vspace{-20pt}
\caption{Number of blocks stacked vs fractions of random training data.}
\label{tab:suboptimal_blocks}
\begin{tabular}{l|c|c|c} 
\toprule
\% Random Data  & 0\% & 50\% & 75\%\\
\midrule
\# Blocks Stacked & 3.5 & 2.0 & 1.0\\
\bottomrule
\end{tabular}
\end{wraptable}

We investigate the influence of the training data on the performance of our approach. In particular, we test whether it is possible to learn effective skill priors from heavily sub-optimal data.

For the experiments in Section~\ref{sec:experiments} we trained the skill prior from high-quality experience collected using expert-like policies, albeit on tasks that differ from the downstream task (see Section~\ref{sec:env_data_details}). However, in many practical cases the quality of training data can be mixed. In this Section we investigate two scenarios for training from sub-optimal data.

\paragraph{Mixed Expert and Random Data.} We assume that a large fraction of the data is collected by inexperienced users leading to very low quality trajectories, while another part of the data is collected by experts and has high quality. We emulate this in the block stacking environment by combining rollouts collected by executing random actions with parts of the high-quality rollouts we used for the experiments in Section~\ref{sec:experiments_blocks_kitchen}.

The results are shown in table~\ref{tab:suboptimal_blocks}. Our approach achieves good performance even when half of the training data consists of very low quality rollouts, and can learn meaningful skills for stacking blocks when \SI{75}{\percent} of the data is of low quality. The best baseline was only able to stack 1.5 blocks on average, even though it was trained from only high-quality data (see Fig.~\ref{fig:training_curves}, middle).

\begin{wrapfigure}{R}{0.4\textwidth}
    \centering
    \vspace{-15pt}
    \includegraphics[width=1\linewidth]{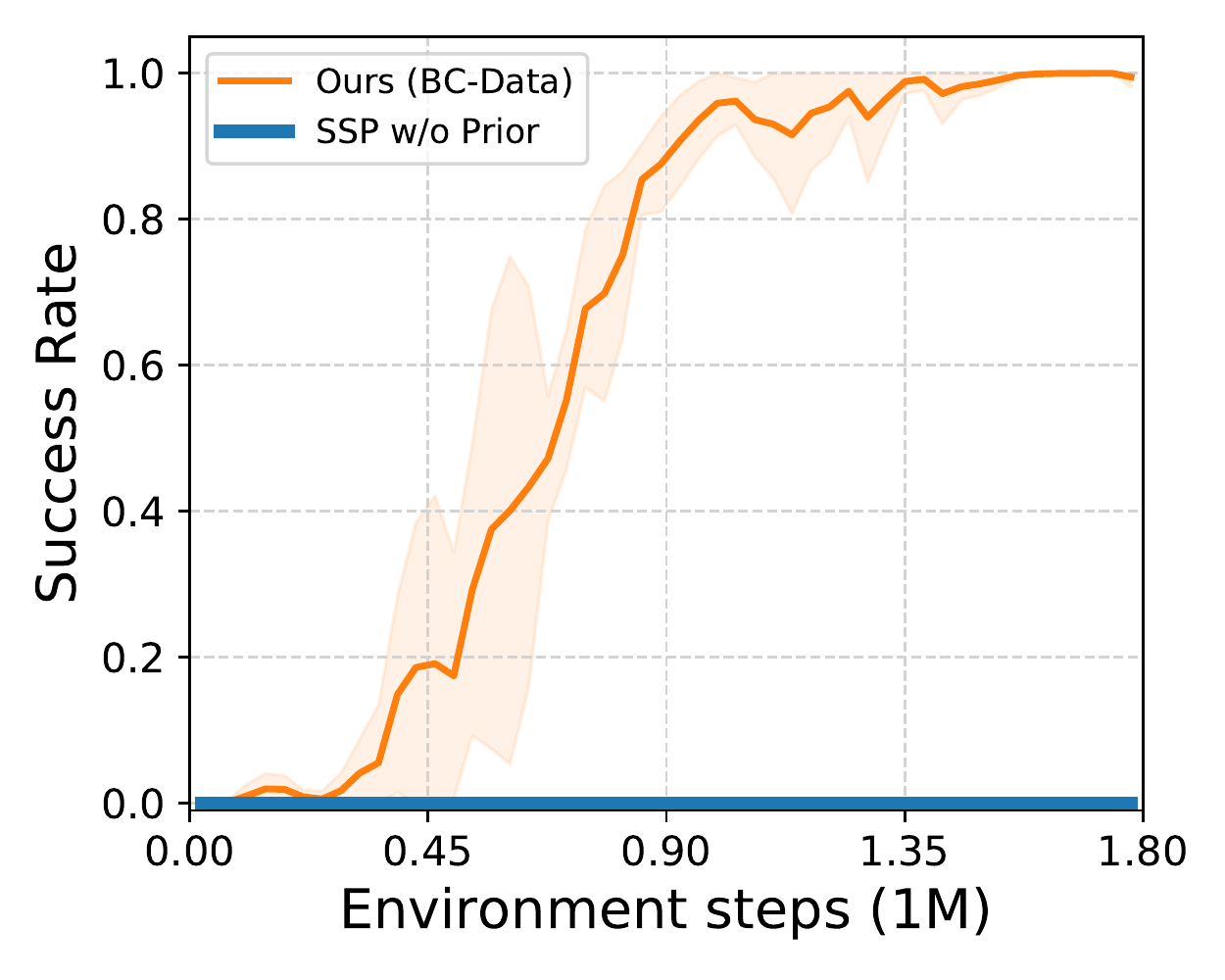}
    \vspace{-15pt}
    \caption{Success rate on maze environment with sub-optimal training data. Our approach, using a prior learned from sub-optimal data generated with the BC policy, is able to reliably learn to reach the goal while the baseline that does not use the learned prior fails.
    }
    \vspace{-10pt}
    \label{fig:bc_data}
\end{wrapfigure}

\paragraph{Only Non-Expert Data.} We assume access to a dataset of only mediocre quality demonstrations, without any expert-level trajectories. We generate this dataset in the maze environment by training a behavior cloning (BC) policy on expert-level trajectories and using it to collect a new dataset. Due to the limited capacity of the BC policy, this dataset is of substantially lower quality than the expert dataset, e.g. the agent collides with walls on average 2.5 times per trajectory while it never collides in expert trajectories.

While we find that a skill-prior-regularized agent trained on the mediocre data explores the maze less widely than one trained on the expert data, it still works substantially better than the baseline that does not use the skill prior, achieving \SI{100}{\percent} success rate of reaching a faraway goal after <1M environment steps, while the baseline does not reach the goal even after 3M environment steps.

Both scenarios show that our approach can learn effective skill embeddings and skill priors even from substantially sub-optimal data.

\section{Reuse of Learned Skill Priors}
\label{sec:prior_reuse}

\begin{figure}
    \centering
    \includegraphics[width=1\linewidth]{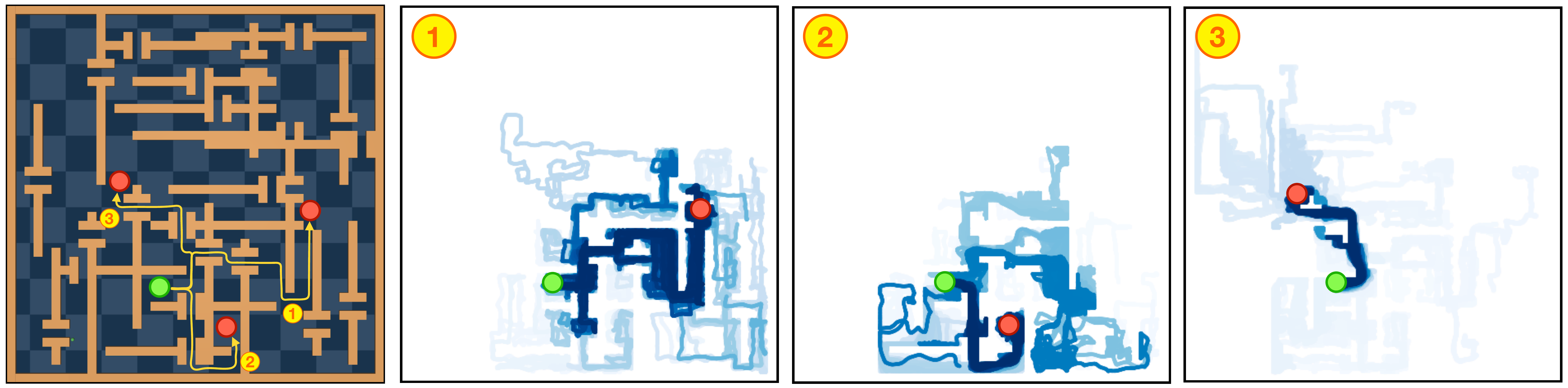}
    \caption{
    Reuse of one learned skill prior for multiple downstream tasks. We train a single skill embedding and skill prior model and then use it to guide downstream RL for multiple tasks. \textbf{Left}: We test prior reuse on three different maze navigation tasks in the form of different goals that need to be reached. \textbf{(1)-(3)}: Agent rollouts during training; the darker the rollout paths, the later during training they were collected. The same prior enables efficient exploration for all three tasks, but allows for convergence to task-specific policies that reach each of the goals upon convergence.
    }
    \label{fig:multi_goal}
\end{figure}

Our approach has two separate stages: (1)~learning of skill embedding and skill prior from offline data and (2)~prior-regularized downstream RL. Since the learning of the skill prior is \emph{independent} of the downstream task, we can reuse the same skill prior for guiding learning on multiple downstream tasks. To test this, we learn a single skill prior on the maze environment depicted in Fig.~\ref{fig:env_overview}~(left) and use it to train multiple downstream task agents that reach different goals.

In Fig.~\ref{fig:multi_goal} we show a visualization of the training rollouts in a top-down view, similar to the visualization in Fig.~\ref{fig:exploration}; darker trajectories are more recent. We can see that the same prior is able to guide downstream agents to efficiently learn to reach diverse goals. All agents achieve $\sim \SI{100}{\percent}$ success rate upon convergence. Intuitively, the prior captures the knowledge that it is more meaningful to e.g. cross doorways instead of bumping into walls, which helps exploration in the maze independent of the goal position.

%% file: main.bbl
\begin{thebibliography}{49}
\providecommand{\natexlab}[1]{#1}
\providecommand{\url}[1]{\texttt{#1}}
\expandafter\ifx\csname urlstyle\endcsname\relax
  \providecommand{\doi}[1]{doi: #1}\else
  \providecommand{\doi}{doi: \begingroup \urlstyle{rm}\Url}\fi

\bibitem[Woodworth and Thorndike(1901)]{woodworth1901influence}
R.~S. Woodworth and E.~Thorndike.
\newblock The influence of improvement in one mental function upon the
  efficiency of other functions.(i).
\newblock \emph{Psychological review}, 8\penalty0 (3):\penalty0 247, 1901.

\bibitem[Caesar et~al.(2019)Caesar, Bankiti, Lang, Vora, Liong, Xu, Krishnan,
  Pan, Baldan, and Beijbom]{nuscenes2019}
H.~Caesar, V.~Bankiti, A.~H. Lang, S.~Vora, V.~E. Liong, Q.~Xu, A.~Krishnan,
  Y.~Pan, G.~Baldan, and O.~Beijbom.
\newblock nuscenes: A multimodal dataset for autonomous driving.
\newblock \emph{preprint arXiv:1903.11027}, 2019.

\bibitem[Mo et~al.(2018)Mo, Li, Lin, and Lee]{Mo18AdobeIndoorNav}
K.~Mo, H.~Li, Z.~Lin, and J.-Y. Lee.
\newblock {The AdobeIndoorNav Dataset}: Towards deep reinforcement learning
  based real-world indoor robot visual navigation.
\newblock \emph{preprint arXiv:1802.08824}, 2018.

\bibitem[Dasari et~al.(2019)Dasari, Ebert, Tian, Nair, Bucher, Schmeckpeper,
  Singh, Levine, and Finn]{dasari2019robonet}
S.~Dasari, F.~Ebert, S.~Tian, S.~Nair, B.~Bucher, K.~Schmeckpeper, S.~Singh,
  S.~Levine, and C.~Finn.
\newblock Robonet: Large-scale multi-robot learning.
\newblock \emph{CoRL}, 2019.

\bibitem[Cabi et~al.(2019)Cabi, Colmenarejo, Novikov, Konyushkova, Reed, Jeong,
  Zolna, Aytar, Budden, Vecerik, Sushkov, Barker, Scholz, Denil, de~Freitas,
  and Wang]{cabi2019}
S.~Cabi, S.~G. Colmenarejo, A.~Novikov, K.~Konyushkova, S.~Reed, R.~Jeong,
  K.~Zolna, Y.~Aytar, D.~Budden, M.~Vecerik, O.~Sushkov, D.~Barker, J.~Scholz,
  M.~Denil, N.~de~Freitas, and Z.~Wang.
\newblock Scaling data-driven robotics with reward sketching and batch
  reinforcement learning.
\newblock \emph{RSS}, 2019.

\bibitem[Schaal(2006)]{schaal2006adaptive}
S.~Schaal.
\newblock \emph{Dynamic Movement Primitives - A Framework for Motor Control in
  Humans and Humanoid Robotics}.
\newblock Springer Tokyo, 2006.

\bibitem[Merel et~al.(2019)Merel, Hasenclever, Galashov, Ahuja, Pham, Wayne,
  Teh, and Heess]{merel2018neural}
J.~Merel, L.~Hasenclever, A.~Galashov, A.~Ahuja, V.~Pham, G.~Wayne, Y.~W. Teh,
  and N.~Heess.
\newblock Neural probabilistic motor primitives for humanoid control.
\newblock \emph{ICLR}, 2019.

\bibitem[Merel et~al.(2020)Merel, Tunyasuvunakool, Ahuja, Tassa, Hasenclever,
  Pham, Erez, Wayne, and Heess]{merel2019reusable}
J.~Merel, S.~Tunyasuvunakool, A.~Ahuja, Y.~Tassa, L.~Hasenclever, V.~Pham,
  T.~Erez, G.~Wayne, and N.~Heess.
\newblock Catch \& carry: Reusable neural controllers for vision-guided
  whole-body tasks.
\newblock \emph{ACM. Trans. Graph.}, 2020.

\bibitem[Shankar et~al.(2019)Shankar, Tulsiani, Pinto, and
  Gupta]{shankar2019discovering}
T.~Shankar, S.~Tulsiani, L.~Pinto, and A.~Gupta.
\newblock Discovering motor programs by recomposing demonstrations.
\newblock In \emph{International Conference on Learning Representations}, 2019.

\bibitem[Lynch et~al.(2020)Lynch, Khansari, Xiao, Kumar, Tompson, Levine, and
  Sermanet]{lynch2020learning}
C.~Lynch, M.~Khansari, T.~Xiao, V.~Kumar, J.~Tompson, S.~Levine, and
  P.~Sermanet.
\newblock Learning latent plans from play.
\newblock In \emph{Conference on Robot Learning}, pages 1113--1132, 2020.

\bibitem[Hausman et~al.(2018)Hausman, Springenberg, Wang, Heess, and
  Riedmiller]{hausman2018learning}
K.~Hausman, J.~T. Springenberg, Z.~Wang, N.~Heess, and M.~Riedmiller.
\newblock Learning an embedding space for transferable robot skills.
\newblock In \emph{International Conference on Learning Representations}, 2018.

\bibitem[Sharma et~al.(2020)Sharma, Gu, Levine, Kumar, and
  Hausman]{sharma2019dynamics}
A.~Sharma, S.~Gu, S.~Levine, V.~Kumar, and K.~Hausman.
\newblock Dynamics-aware unsupervised discovery of skills.
\newblock \emph{ICLR}, 2020.

\bibitem[Jong et~al.(2008)Jong, Hester, and Stone]{jong2008utility}
N.~K. Jong, T.~Hester, and P.~Stone.
\newblock The utility of temporal abstraction in reinforcement learning.
\newblock In \emph{AAMAS (1)}, pages 299--306. Citeseer, 2008.

\bibitem[Finn et~al.(2017)Finn, Abbeel, and Levine]{finn2017model}
C.~Finn, P.~Abbeel, and S.~Levine.
\newblock Model-agnostic meta-learning for fast adaptation of deep networks.
\newblock \emph{ICML}, 2017.

\bibitem[Rakelly et~al.(2019)Rakelly, Zhou, Finn, Levine, and
  Quillen]{rakelly2019efficient}
K.~Rakelly, A.~Zhou, C.~Finn, S.~Levine, and D.~Quillen.
\newblock Efficient off-policy meta-reinforcement learning via probabilistic
  context variables.
\newblock In \emph{ICML}, 2019.

\bibitem[Levine et~al.(2020)Levine, Kumar, Tucker, and Fu]{levine2020offline}
S.~Levine, A.~Kumar, G.~Tucker, and J.~Fu.
\newblock Offline reinforcement learning: Tutorial, review, and perspectives on
  open problems.
\newblock \emph{arXiv preprint arXiv:2005.01643}, 2020.

\bibitem[Fujimoto et~al.(2019)Fujimoto, Meger, and Precup]{fujimoto2019off}
S.~Fujimoto, D.~Meger, and D.~Precup.
\newblock Off-policy deep reinforcement learning without exploration.
\newblock In \emph{International Conference on Machine Learning}, pages
  2052--2062, 2019.

\bibitem[Jaques et~al.(2019)Jaques, Ghandeharioun, Shen, Ferguson, Lapedriza,
  Jones, Gu, and Picard]{jaques2019way}
N.~Jaques, A.~Ghandeharioun, J.~H. Shen, C.~Ferguson, A.~Lapedriza, N.~Jones,
  S.~Gu, and R.~Picard.
\newblock Way off-policy batch deep reinforcement learning of implicit human
  preferences in dialog.
\newblock \emph{arXiv preprint arXiv:1907.00456}, 2019.

\bibitem[Kumar et~al.(2019)Kumar, Fu, Soh, Tucker, and
  Levine]{kumar2019stabilizing}
A.~Kumar, J.~Fu, M.~Soh, G.~Tucker, and S.~Levine.
\newblock Stabilizing off-policy q-learning via bootstrapping error reduction.
\newblock In \emph{Advances in Neural Information Processing Systems}, pages
  11784--11794, 2019.

\bibitem[Wu et~al.(2019)Wu, Tucker, and Nachum]{wu2019behavior}
Y.~Wu, G.~Tucker, and O.~Nachum.
\newblock Behavior regularized offline reinforcement learning.
\newblock \emph{arXiv preprint arXiv:1911.11361}, 2019.

\bibitem[Nair et~al.(2020)Nair, Dalal, Gupta, and Levine]{nair2020accelerating}
A.~Nair, M.~Dalal, A.~Gupta, and S.~Levine.
\newblock Accelerating online reinforcement learning with offline datasets.
\newblock \emph{arXiv preprint arXiv:2006.09359}, 2020.

\bibitem[Taylor and Stone(2009)]{taylor2009transfer}
M.~E. Taylor and P.~Stone.
\newblock Transfer learning for reinforcement learning domains: A survey.
\newblock \emph{Journal of Machine Learning Research}, 10\penalty0 (7), 2009.

\bibitem[Thrun and Schwartz(1995)]{thrun1995finding}
S.~Thrun and A.~Schwartz.
\newblock Finding structure in reinforcement learning.
\newblock In \emph{NIPS}, 1995.

\bibitem[Pickett and Barto(2002)]{pickett2002policyblocks}
M.~Pickett and A.~G. Barto.
\newblock Policyblocks: An algorithm for creating useful macro-actions in
  reinforcement learning.
\newblock In \emph{ICML}, volume~19, pages 506--513, 2002.

\bibitem[Sutton et~al.(1999)Sutton, Precup, and Singh]{sutton1999between}
R.~S. Sutton, D.~Precup, and S.~Singh.
\newblock Between mdps and semi-mdps: A framework for temporal abstraction in
  reinforcement learning.
\newblock \emph{Artificial intelligence}, 112\penalty0 (1-2):\penalty0
  181--211, 1999.

\bibitem[Bacon et~al.(2017)Bacon, Harb, and Precup]{bacon2017option}
P.-L. Bacon, J.~Harb, and D.~Precup.
\newblock The option-critic architecture.
\newblock In \emph{AAAI}, 2017.

\bibitem[Gupta et~al.(2019)Gupta, Kumar, Lynch, Levine, and
  Hausman]{gupta2019relay}
A.~Gupta, V.~Kumar, C.~Lynch, S.~Levine, and K.~Hausman.
\newblock Relay policy learning: Solving long-horizon tasks via imitation and
  reinforcement learning.
\newblock \emph{CoRL}, 2019.

\bibitem[Mandlekar et~al.(2020)Mandlekar, Ramos, Boots, Fei-Fei, Garg, and
  Fox]{mandlekar2019iris}
A.~Mandlekar, F.~Ramos, B.~Boots, L.~Fei-Fei, A.~Garg, and D.~Fox.
\newblock Iris: Implicit reinforcement without interaction at scale for
  learning control from offline robot manipulation data.
\newblock \emph{ICRA}, 2020.

\bibitem[Lee et~al.(2018)Lee, Sun, Somasundaram, Hu, and Lim]{lee2018composing}
Y.~Lee, S.-H. Sun, S.~Somasundaram, E.~S. Hu, and J.~J. Lim.
\newblock Composing complex skills by learning transition policies.
\newblock In \emph{International Conference on Learning Representations}, 2018.

\bibitem[Kipf et~al.(2019)Kipf, Li, Dai, Zambaldi, Grefenstette, Kohli, and
  Battaglia]{kipf2018compositional}
T.~Kipf, Y.~Li, H.~Dai, V.~Zambaldi, E.~Grefenstette, P.~Kohli, and
  P.~Battaglia.
\newblock Compositional imitation learning: Explaining and executing one task
  at a time.
\newblock \emph{ICML}, 2019.

\bibitem[Whitney et~al.(2020)Whitney, Agarwal, Cho, and
  Gupta]{whitney2019dynamics}
W.~Whitney, R.~Agarwal, K.~Cho, and A.~Gupta.
\newblock Dynamics-aware embeddings.
\newblock \emph{ICLR}, 2020.

\bibitem[Siegel et~al.(2020)Siegel, Springenberg, Berkenkamp, Abdolmaleki,
  Neunert, Lampe, Hafner, and Riedmiller]{siegel2020keep}
N.~Y. Siegel, J.~T. Springenberg, F.~Berkenkamp, A.~Abdolmaleki, M.~Neunert,
  T.~Lampe, R.~Hafner, and M.~Riedmiller.
\newblock Keep doing what worked: Behavioral modelling priors for offline
  reinforcement learning.
\newblock \emph{ICLR}, 2020.

\bibitem[Fu et~al.(2020)Fu, Kumar, Nachum, Tucker, and Levine]{fu2020d4rl}
J.~Fu, A.~Kumar, O.~Nachum, G.~Tucker, and S.~Levine.
\newblock D4rl: Datasets for deep data-driven reinforcement learning.
\newblock \emph{arXiv preprint arXiv:2004.07219}, 2020.

\bibitem[Gulcehre et~al.(2020)Gulcehre, Wang, Novikov, Paine, Colmenarejo,
  Zolna, Agarwal, Merel, Mankowitz, Paduraru, et~al.]{gulcehre2020rl}
C.~Gulcehre, Z.~Wang, A.~Novikov, T.~L. Paine, S.~G. Colmenarejo, K.~Zolna,
  R.~Agarwal, J.~Merel, D.~Mankowitz, C.~Paduraru, et~al.
\newblock Rl unplugged: Benchmarks for offline reinforcement learning.
\newblock \emph{arXiv preprint arXiv:2006.13888}, 2020.

\bibitem[Schaal et~al.(2005)Schaal, Peters, Nakanishi, and
  Ijspeert]{schaal2005motion}
S.~Schaal, J.~Peters, J.~Nakanishi, and A.~Ijspeert.
\newblock Learning movement primitives.
\newblock In P.~Dario and R.~Chatila, editors, \emph{Robotics Research}.
  Springer Berlin Heidelberg, 2005.

\bibitem[Mandlekar et~al.(2018)Mandlekar, Zhu, Garg, Booher, Spero, Tung, Gao,
  Emmons, Gupta, Orbay, Savarese, and Fei-Fei]{mandlekar2018roboturk}
A.~Mandlekar, Y.~Zhu, A.~Garg, J.~Booher, M.~Spero, A.~Tung, J.~Gao, J.~Emmons,
  A.~Gupta, E.~Orbay, S.~Savarese, and L.~Fei-Fei.
\newblock Roboturk: A crowdsourcing platform for robotic skill learning through
  imitation.
\newblock In \emph{Conference on Robot Learning}, 2018.

\bibitem[Fang et~al.(2019)Fang, Zhu, Garg, Savarese, and
  Fei-Fei]{fang2019dynamics}
K.~Fang, Y.~Zhu, A.~Garg, S.~Savarese, and L.~Fei-Fei.
\newblock Dynamics learning with cascaded variational inference for multi-step
  manipulation.
\newblock \emph{CoRL 2019}, 2019.

\bibitem[Pertsch et~al.(2020)Pertsch, Rybkin, Yang, Zhou, Derpanis, Lim,
  Daniilidis, and Jaegle]{pertsch2020keyin}
K.~Pertsch, O.~Rybkin, J.~Yang, S.~Zhou, K.~Derpanis, J.~Lim, K.~Daniilidis,
  and A.~Jaegle.
\newblock Keyin: Keyframing for visual planning.
\newblock \emph{Conference on Learning for Dynamics and Control}, 2020.

\bibitem[Higgins et~al.(2017)Higgins, Matthey, Pal, Burgess, Glorot, Botvinick,
  Mohamed, and Lerchner]{higgins2017beta}
I.~Higgins, L.~Matthey, A.~Pal, C.~Burgess, X.~Glorot, M.~Botvinick,
  S.~Mohamed, and A.~Lerchner.
\newblock beta-{VAE}: Learning basic visual concepts with a constrained
  variational framework.
\newblock In \emph{ICLR}, 2017.

\bibitem[Kingma and Welling(2014)]{kingma2014auto}
D.~P. Kingma and M.~Welling.
\newblock Auto-encoding variational {B}ayes.
\newblock In \emph{ICLR}, 2014.

\bibitem[Rezende et~al.(2014)Rezende, Mohamed, and
  Wierstra]{rezende2014stochastic}
D.~J. Rezende, S.~Mohamed, and D.~Wierstra.
\newblock Stochastic backpropagation and approximate inference in deep
  generative models.
\newblock In \emph{ICML}, 2014.

\bibitem[Bishop(2006)]{bishop2006pattern}
C.~M. Bishop.
\newblock \emph{Pattern recognition and machine learning}.
\newblock springer, 2006.

\bibitem[Rezende and Mohamed(2015)]{rezende2015variational}
D.~J. Rezende and S.~Mohamed.
\newblock Variational inference with normalizing flows.
\newblock \emph{ICML}, 2015.

\bibitem[Dinh et~al.(2017)Dinh, Sohl-Dickstein, and Bengio]{dinh2016density}
L.~Dinh, J.~Sohl-Dickstein, and S.~Bengio.
\newblock Density estimation using real nvp.
\newblock \emph{ICLR}, 2017.

\bibitem[Ziebart(2010)]{ziebart2010modeling}
B.~D. Ziebart.
\newblock Modeling purposeful adaptive behavior with the principle of maximum
  causal entropy.
\newblock 2010.

\bibitem[Levine(2018)]{levine2018reinforcement}
S.~Levine.
\newblock Reinforcement learning and control as probabilistic inference:
  Tutorial and review.
\newblock \emph{arXiv preprint arXiv:1805.00909}, 2018.

\bibitem[Haarnoja et~al.(2018{\natexlab{a}})Haarnoja, Zhou, Abbeel, and
  Levine]{haarnoja2018sac}
T.~Haarnoja, A.~Zhou, P.~Abbeel, and S.~Levine.
\newblock Soft actor-critic: Off-policy maximum entropy deep reinforcement
  learning with a stochastic actor.
\newblock \emph{ICML}, 2018{\natexlab{a}}.

\bibitem[Haarnoja et~al.(2018{\natexlab{b}})Haarnoja, Zhou, Hartikainen,
  Tucker, Ha, Tan, Kumar, Zhu, Gupta, Abbeel,
  et~al.]{haarnoja2018sac_algo_applications}
T.~Haarnoja, A.~Zhou, K.~Hartikainen, G.~Tucker, S.~Ha, J.~Tan, V.~Kumar,
  H.~Zhu, A.~Gupta, P.~Abbeel, et~al.
\newblock Soft actor-critic algorithms and applications.
\newblock \emph{arXiv preprint arXiv:1812.05905}, 2018{\natexlab{b}}.

\bibitem[Sutton and Barto(2018)]{sutton2018reinforcement}
R.~S. Sutton and A.~G. Barto.
\newblock \emph{Reinforcement learning: An introduction}.
\newblock MIT press, 2018.

\end{thebibliography}
